\pdfoutput=1

\documentclass[11pt]{article}

\usepackage[]{EMNLP2025}

\usepackage{times}
\usepackage{latexsym}
\usepackage{enumitem} 
\usepackage{CJKutf8}
\usepackage{booktabs}
\usepackage{times}
\usepackage{hyperref}
\usepackage{latexsym}
\usepackage{booktabs}
\usepackage{algorithm}
\usepackage{algorithmic}
\usepackage{multicol}
\usepackage{amsfonts}
\usepackage{graphicx}
\usepackage{booktabs}
\usepackage[table]{xcolor}
\usepackage{multirow}
\usepackage{makecell} 
\usepackage{array} 
\usepackage{fontawesome} 
\usepackage{xcolor} 
\usepackage{caption} 
\usepackage{amssymb} 
\usepackage{amsmath} 
\usepackage{colortbl}
\usepackage[listings,skins,breakable]{tcolorbox}
\usepackage{pifont}
\usepackage{subcaption} 
\usepackage{xspace}
\usepackage{float} 

\newcommand{\halfcorrect}{\textcolor{blue}{\ding{52}\rotatebox[origin=c]{-9.2}{\kern-0.7em\ding{55}}}}
\definecolor{myrowcolor}{RGB}{253,249,234}

\usepackage[T1]{fontenc}

\usepackage[utf8]{inputenc}

\usepackage{microtype}

\usepackage{inconsolata}

\title{DecoupledESC: Enhancing Emotional Support Generation via Strategy-Response Decoupled Preference Optimization}

\author{Chao Zhang, Xin Shi, Xueqiao Zhang, Yifan Zhu, Yi Yang, Yawei Luo\thanks{\ \ Corresponding author}\\
  Zhejiang University \\
  \texttt{\{chao\_zhang, yaweiluo\}@zju.edu.cn}}

\begin{document}
\maketitle
\begin{abstract}
Recent advances in Emotional Support Conversation (ESC) have improved emotional support generation by fine-tuning Large Language Models (LLMs) via Supervised Fine-Tuning (SFT). However, common psychological errors still persist. While Direct Preference Optimization (DPO) shows promise in reducing such errors through pairwise preference learning, its effectiveness in ESC tasks is limited by two key challenges: \textbf{(1) Entangled data structure:} Existing ESC data inherently entangles psychological strategies and response content, making it difficult to construct high-quality preference pairs; and \textbf{(2) Optimization ambiguity:} Applying vanilla DPO to such entangled pairwise data leads to ambiguous training objectives. To address these issues, we introduce Inferential Preference Mining (IPM) to construct high-quality preference data, forming the \texttt{IPM-PrefDial} dataset. Building upon this data, we propose a \textbf{Decoupled ESC} framework inspired by Gross’s Extended Process Model of Emotion Regulation, which decomposes the ESC task into two sequential subtasks: strategy planning and empathic response generation. Each was trained via SFT and subsequently enhanced by DPO to align with the psychological preference. Extensive experiments demonstrate that our Decoupled ESC framework outperforms baselines, reducing preference bias and improving response quality\footnote{Our data and code are available at \url{https://github.com/Zc0812/DecoupledESC}.}.

\end{abstract}


\begin{figure}[t]
    \centering
    \includegraphics[width=1\linewidth, trim=0 112 360 0, clip]{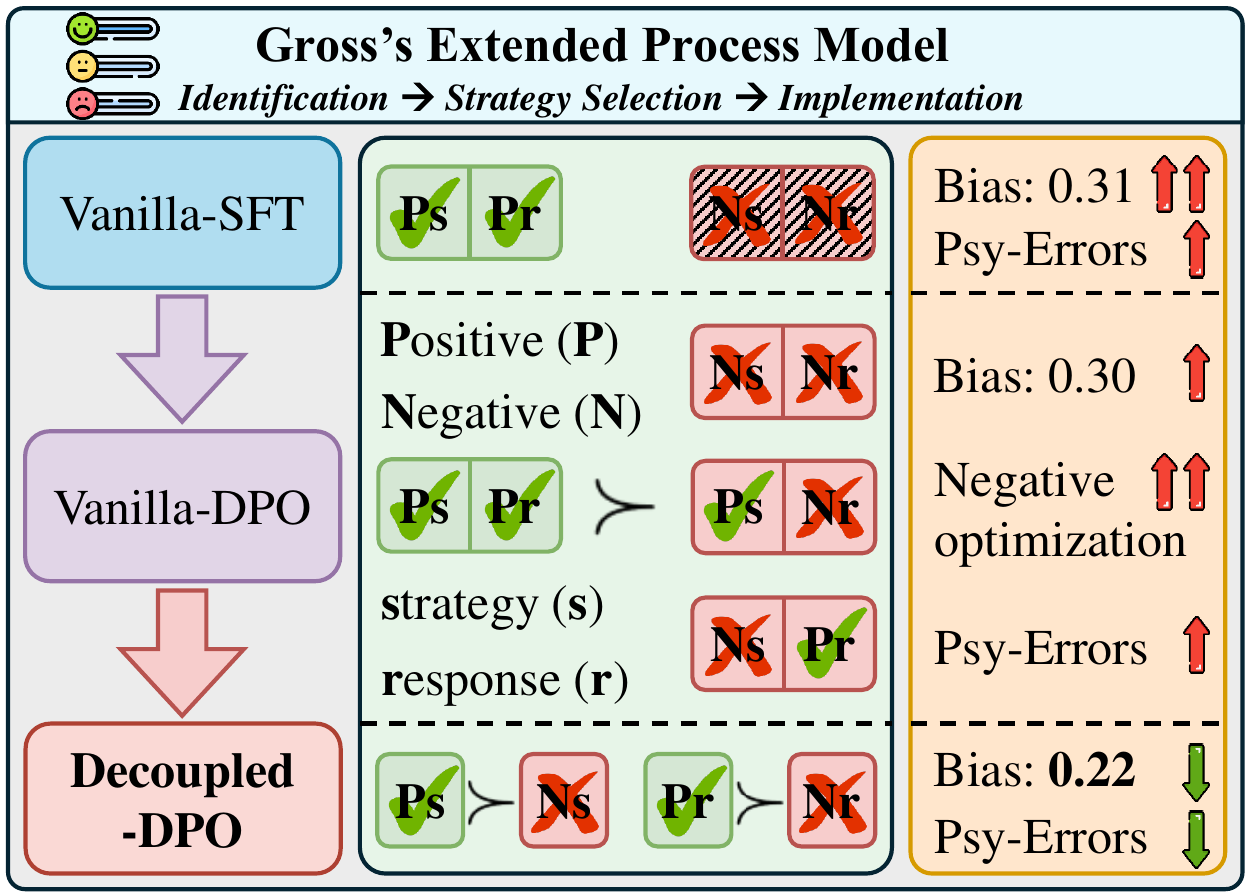}
    \caption{Comparison from Vanilla-SFT to Vanilla-DPO to Decoupled-DPO. Vanilla-SFT lacks negative preference data, leading to high preference bias; Vanilla-DPO uses coupled preference data, causing potential negative optimization (regards PsNr, NsPr as pure negative samples); Decoupled-DPO decouples strategy and response, effectively reducing bias and psychological errors.}
    \label{fig: Introduction}
\end{figure}

\section{Introduction}
Mental health is essential to well-being~\cite{prince2007no}, yet rising stress and fast-paced life have increased related issues~\cite{bor2014child, brundtland2000mental, paisley2001school, ma2023personas}. According to WHO, 1/8 people suffer from mental disorders~\cite{world2022world}. Amid a shortage of professionals, this underscores the need for scalable solutions, where Large Language Models (LLMs) offer promising potential.

To enhance the performance of LLMs in Emotional Support Conversation (ESC), prior works~\cite{zhang2024cpsycoun, chen2023soulchat} have constructed several large-scale, high-quality dialogue datasets and applied Supervised Fine-Tuning (SFT) to improve model responses. Among them, Liu \textit{et al.}~\cite{liu2021towards} built the \texttt{ESConv} dataset based on Hill’s Helping Skills Theory~\cite{hill1999helping} and filtered out \texttt{FailedESConv} dataset. The \texttt{ESConv} dataset follows a three-phase structure (Exploration → Comfort → Action) and includes eight types of support strategies, each paired with corresponding responses, details are provided in Appendix~\ref{sec: Definition} and~\ref{sec: ESConv}. This structured design significantly enhances a model’s ability to generate empathetic dialogue.

\textbf{Observation}\quad Currently, SFT has become the mainstream approach in the ESC field. However, we observe that models still frequently exhibit common psychological errors~\cite{raskin2005person, stebnicki2007empathy} during inference, which align with those identified in the \texttt{FailedESConv} dataset (\hyperref[sec: obs1]{Obs 1}). In addition, Zhao \textit{et al.}~\cite{zhao2025chain} found that SFT's focus on single gold strategy-response pairs limits adaptability to nuanced contexts, weakening empathetic support. To mitigate this, they use Monte Carlo Tree Search (MCTS) to collect pairwise preference data linking strategies and responses, and apply Direct Preference Optimization (Vanilla-DPO) to guide the model in choosing appropriate strategies, thereby partially reducing preference bias and improving response quality.

\textbf{Challenges}\quad However, as shown in Figure~\ref{fig: Introduction} and ~\ref{fig: Preliminary_3}, our further analysis reveals that the limitations of current work lie not in the SFT or DPO training methods themselves, but rather in two overlooked challenges (\hyperref[sec: obs2]{Obs 2}): \textbf{(1) Entangled data structure:} Existing ESC datasets heavily entangle psychological strategies with response content, making it difficult to construct high-quality preference pairs. For instance, penalizing responses with correct strategies but flawed content may degrade data quality.
\textbf{(2) Optimization ambiguity:} Applying Vanilla-DPO directly to such entangled data can blur training objectives and even lead to negative optimization outcomes.

\textbf{Approach}\quad To address these issues, we first introduce the Inferential Preference Mining (IPM) method, which automatically constructs preference samples decoupled from strategy-response. Specifically, we use dynamic data routing to route four types of psychological error samples identified from the SFT model's inference data to the DPO training stage of either strategy planning or response generation, depending on the error type. These samples are then paired with human-annotated ground truth samples to form the \textbf{\underline{I}}nferential \textbf{\underline{P}}reference \textbf{\underline{M}}ining \textbf{\underline{Pref}}erence \textbf{\underline{Dial}}ogues (\texttt{IPM-PrefDial}) dataset, containing $21$k strategy preference pairs and $11$k response preference pairs. This dataset provides disentangled and high-quality supervision signals for two separate DPO models. Building on this, we propose a \textbf{Decoupled ESC optimization framework (DecoupledESC)}, grounded in the Extended Process Model of Emotion Regulation (EPMER)~\cite{gross2015emotion}, which divides emotion regulation into three sequential stages: (1) Identification → (2) Strategy Selection → (3) Implementation, details are provided in Appendix~\ref{sec: Definition_Gross}. Accordingly, we explicitly split the ESC task into two subtasks: Strategy Planning (SP) and Response Generation (RG).

\textbf{Results}\quad Across multiple evaluation metrics, our decoupled optimization framework significantly outperforms joint training baselines. It not only enhances the diversity of strategy selection but also improves response quality and empathy.

\textbf{Contributions}\quad Our key contributions are summarized as follows:
\begin{itemize} 
    \item We analyze common psychological errors in existing SFT paradigms and introduce Inferential Preference Mining (IPM) and Expert-Guided ICL Annotation to construct \texttt{IPM-PrefDial}, a decoupled strategy–response dataset.

    \item We propose a Decoupled ESC framework, which explicitly splits the ESC task into two subtasks: Strategy Planning and Response Generation, effectively mitigating preference bias and enhancing response quality.

    \item Extensive experiments show that our Decoupled ESC optimization framework significantly outperforms joint optimization baselines across multiple evaluation metrics.
\end{itemize}

\begin{figure*}[t]
    \centering
    \includegraphics[width=1\linewidth, trim=0 212 0 0, clip]{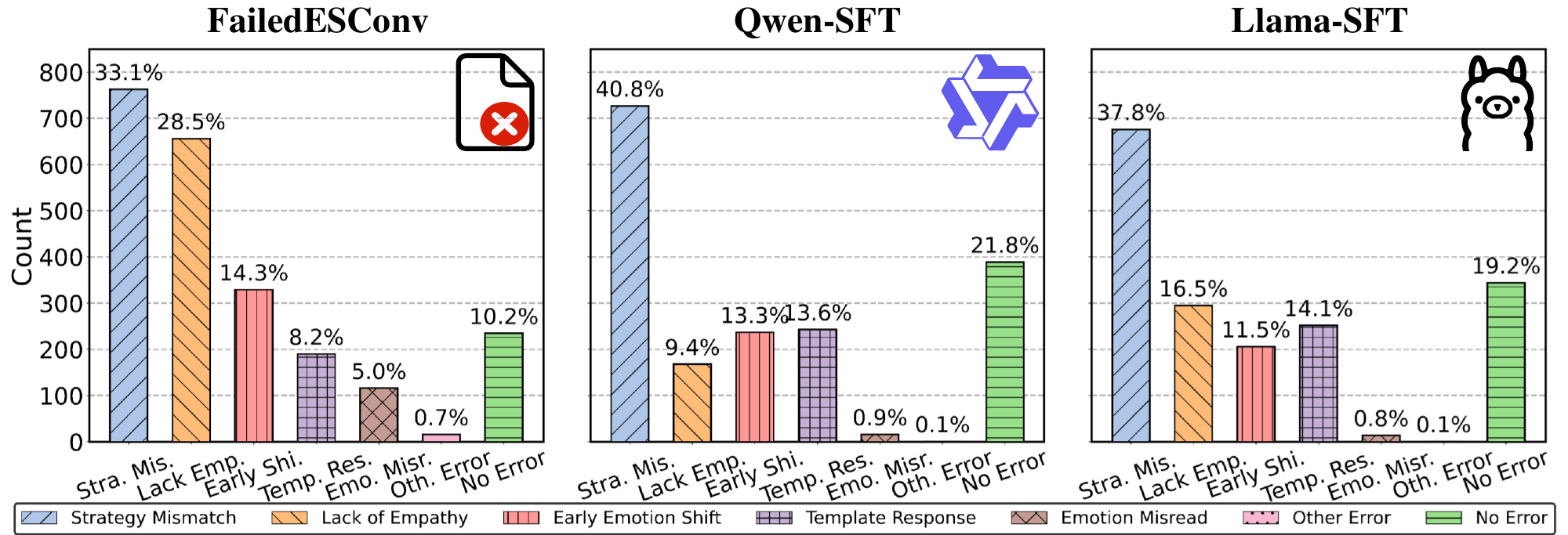}
    \caption{Comparison of common psychological error type proportions among the \texttt{FailedESConv} dataset, Qwen-SFT inference results, and Llama-SFT inference results. \textit{Other Error} refers to non-psychological errors.}
    \label{fig: Preliminary}
    
\end{figure*}

\begin{table*}[t]
\centering
\setlength{\tabcolsep}{4pt}
\resizebox{\textwidth}{!}{%
\begin{tabular}{@{}c ccc ccc ccc ccc ccc ccc ccc ccc @{}}

\toprule[1.2pt]
\multirow{2}{*}{\textbf{Subset}} & \multicolumn{3}{c}{\textbf{Qu}} & \multicolumn{3}{c}{\textbf{RP}} & \multicolumn{3}{c}{\textbf{RF}} & \multicolumn{3}{c}{\textbf{Sd}} & \multicolumn{3}{c}{\textbf{AR}} & \multicolumn{3}{c}{\textbf{PS}} & \multicolumn{3}{c}{\textbf{In}} & \multicolumn{3}{c}{\textbf{Ot}} \\
\cmidrule(lr){2-4} \cmidrule(lr){5-7} \cmidrule(lr){8-10} \cmidrule(lr){11-13} \cmidrule(lr){14-16} \cmidrule(lr){17-19} \cmidrule(lr){20-22} \cmidrule(lr){23-25}
& \textbf{GT} & \textbf{Pred} & \textbf{$\mathcal{B}$} & \textbf{GT} & \textbf{Pred} & \textbf{$\mathcal{B}$} & \textbf{GT} & \textbf{Pred} & \textbf{$\mathcal{B}$} & \textbf{GT} & \textbf{Pred} & \textbf{$\mathcal{B}$} & \textbf{GT} & \textbf{Pred} & \textbf{$\mathcal{B}$} & \textbf{GT} & \textbf{Pred} & \textbf{$\mathcal{B}$} & \textbf{GT} & \textbf{Pred} & \textbf{$\mathcal{B}$} & \textbf{GT} & \textbf{Pred} & \textbf{$\mathcal{B}$} \\
\midrule
\textbf{1} & 110 & 135 & 1.41 & 40 & 25 & 0.57 & 43 & 38 & 0.82 & 39 & 50 & 1.43 & 85 & 50 & 0.55 & 77 & 91 & 1.29 & 26 & 19 & 0.69 & 80 & 92 & 1.25 \\
\textbf{2} & 104 & 122 & 1.32 & 35 & 18 & 0.51 & 41 & 43 & 1.11 & 48 & 42 & 0.90 & 74 & 56 & 0.78 & 73 & 95 & 1.46 & 28 & 20 & 0.76 & 97 & 104 & 1.15 \\
\textbf{3} & 97  & 120 & 1.42 & 29 & 30 & 1.04 & 47 & 37 & 0.79 & 37 & 40 & 1.13 & 79 & 54 & 0.66 & 83 & 95 & 1.24 & 35 & 20 & 0.57 & 93 & 104 & 1.16 \\
\textbf{4} & 93  & 135 & 1.78 & 27 & 13 & 0.45 & 40 & 32 & 0.81 & 40 & 42 & 1.19 & 89 & 49 & 0.51 & 77 & 110 & 1.68 & 32 & 18 & 0.57 & 102 & 101 & 1.02 \\
\bottomrule[1.2pt]
\end{tabular}
}
\caption{Analysis of Strategy Distribution and Preference Bias ($\mathcal{B}$) Across Four Randomly Sampled Subsets. This expanded layout clarifies the relationship between Ground Truth (GT), Predictions (Pred), and the resulting Bias for each strategy (defined in Appendix~\ref{sec: Definition_Strategies}).}
\label{tab: sampling_validation_optimized}
\end{table*}

\section{Related Work}
\paragraph{Emotional Support Conversation.}
Emotional Support Conversation (ESC) aims to alleviate users’ emotional distress through empathetic and supportive responses. Liu \textit{et al.}~\cite{liu2021towards} first introduced the concept and built the \texttt{ESConv} dataset with 8 support strategies, 1.3k dialogues. They also released the \texttt{FailedESConv} dataset, containing 196 failed dialogues. Subsequent studies improved ESC systems by enhancing data quality~\cite{sun-etal-2021-psyqa, qiu-etal-2024-smile, chen2023soulchat}, adding external strategy planners~\cite{deng2024plug, he2024planning, he2025simulation}, and incorporating commonsense reasoning~\cite{tu2022misc, deng2023knowledge, luo2024large}. Supervised Fine-Tuning (SFT) remains the dominant training paradigm with strong real-world performance (e.g., MeChat~\cite{qiu-etal-2024-smile}, SweetieChat~\cite{ye2025sweetiechat}). Recently, preference-based methods like Direct Preference Optimization (DPO)~\cite{rafailov2023direct} have emerged. Zhao \textit{et al.}~\cite{zhao2025chain} introduced DPO with MCTS-based data to jointly optimize strategies and responses. However, the fixed coupling limited independent optimization and resulted in lower response quality.

\paragraph{Reinforcement Learning for LLM.}
Reinforcement Learning (RL) was initially introduced into LLM training to align with human preferences~\cite{ouyang2022training}. This approach uses a reward model to guide the optimization of the policy model via the Proximal Policy Optimization (PPO) algorithm~\cite{schulman2017proximal}. Recently, the Group Relative Policy Optimization (GRPO) algorithm was proposed to enhance model reasoning capabilities~\cite{shao2024deepseekmath}, which eliminates the need for a critic model by using within-group rewards as advantages. While these online reinforcement learning methods are effective, they suffer from high computational costs and reliance on accurate reward modeling. As a simpler offline optimization algorithm, DPO optimizes the policy model from pairwise preference data directly without the need for reward modeling. Due to its simplicity and effectiveness, DPO has achieved significant success across multiple domains, including mathematical reasoning, code generation, and recommendation systems~\cite{lai2024step, zhang-etal-2025-codedpo, zhang2025eduplanner, chen2024softmax}.


\section{Preliminary Observations}
To investigate the causes of low response quality in the ESC task, we analyzed outputs from six models: Base, SFT\footnote{Trained on \texttt{ESConv}~\cite{liu2021towards} datasets. \newline\href{https://github.com/thu-coai/Emotional-Support-Conversation}{github.com/thu-coai/Emotional-Support-Conversation}}, and DPO versions of \texttt{Qwen2.5-7B-Instruct}~\cite{qwen2.5} and \texttt{Llama3.1-8B-Instruct}~\cite{dubey2024llama}.

\subsection{Preference Bias and Psy-Errors (Obs 1)}
\label{sec: obs1}
\begin{figure}[t]
    \centering
    \includegraphics[width=1\linewidth, trim=0 0 30 42, clip]{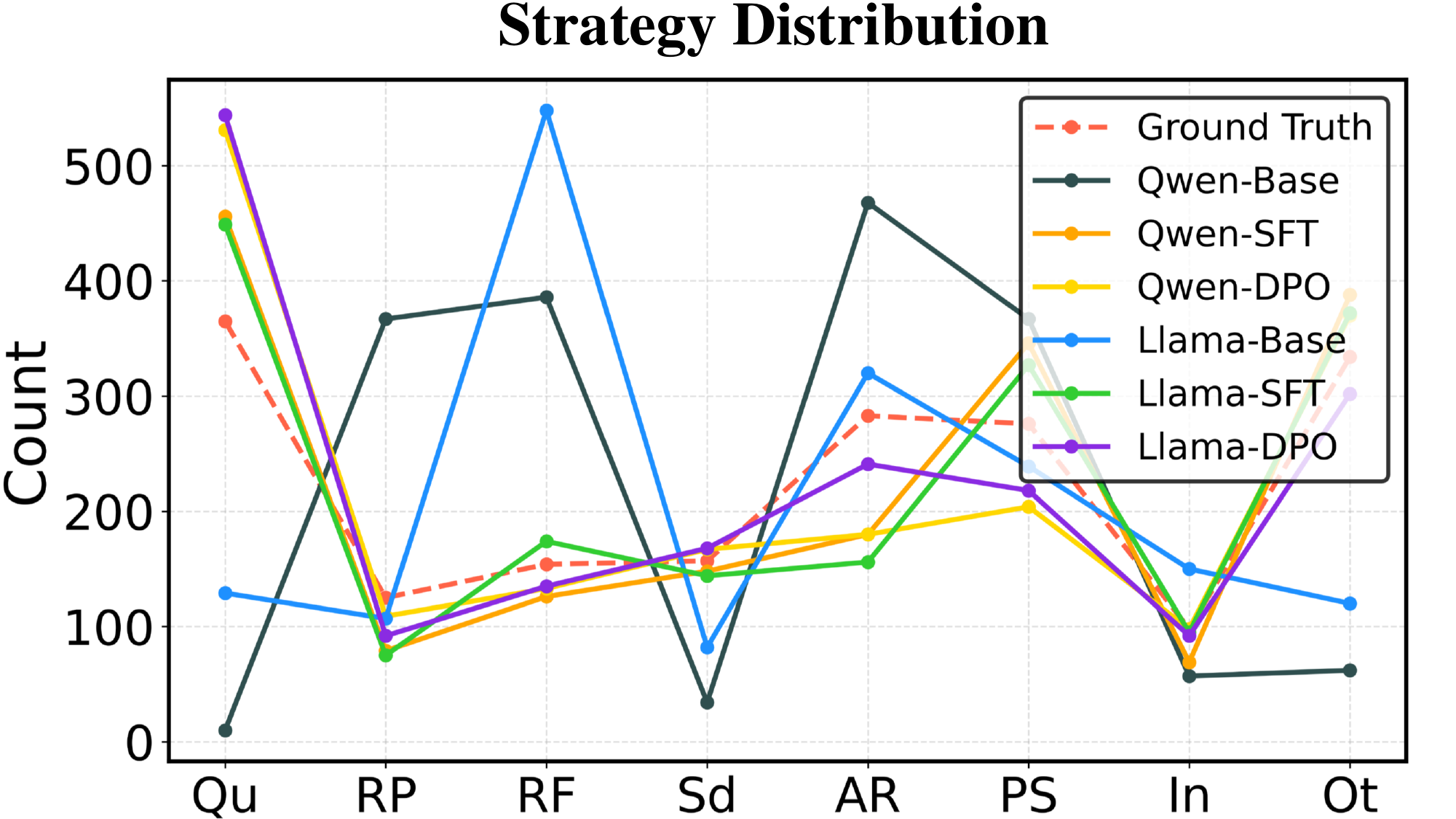}
    \caption{Strategy Distribution across different models.}
    \label{fig: Preliminary_2}
\end{figure}

Current Base and SFT models (e.g., Qwen-Base, Qwen-SFT) show strong strategy preferences~\cite{kang2024can}, often overusing fixed strategies and failing to adapt to users' emotional states. As shown in Figure~\ref{fig: Preliminary_2}, the Base and SFT models show different levels of divergence from the ground truth in strategy distributions. 

Additionally, to verify that the observed strategy preference is an intrinsic model bias, not an artifact of sampling variance, we performed a validation on four random subsets of the test set. As shown in Table~\ref{tab: sampling_validation_optimized}, the results reveal a consistent strategy bias across all subsets. Despite fluctuations in ground truth counts, the preference bias remained stable, strongly suggesting it is an intrinsic characteristic of the model's decision-making process.

To further explore the impact of preference bias on response quality, we compared the outputs of Qwen-Base, Qwen-SFT, Llama-Base, and Llama-SFT with the \texttt{FailedESConv} dataset. As shown in Figure~\ref{fig: Preliminary}, common psychological errors~\cite{raskin2005person, stebnicki2007empathy} observed in the SFT-generated responses frequently aligned with those found in \texttt{FailedESConv}\footnote{We have manually filtered out samples classified as \texttt{FailedESConv} due to non-psychological errors, such as incomplete or repetitive dialogues.}, including: \textit{(1) Strategy Mismatch}, \textit{(2) Lack of Empathy}, \textit{(3) Early Emotion Shift}, \textit{(4) Template Response}, \textit{(5) Emotion Misread}. The definitions and corresponding examples are detailed in Appendix~\ref{sec: PsyError}.

Although SFT reduces some errors, the empathy quality remains unsatisfactory. We argue that this stems from the SFT paradigm's reliance on high-quality samples~\cite{zhao2025chain} without incorporating negative supervision signals from the \texttt{FailedESConv} dataset, failing to address bias in strategy selection and emotional understanding.

\begin{figure}[t]
    \centering
    \includegraphics[width=1\linewidth, trim=25 30 25 45, clip]{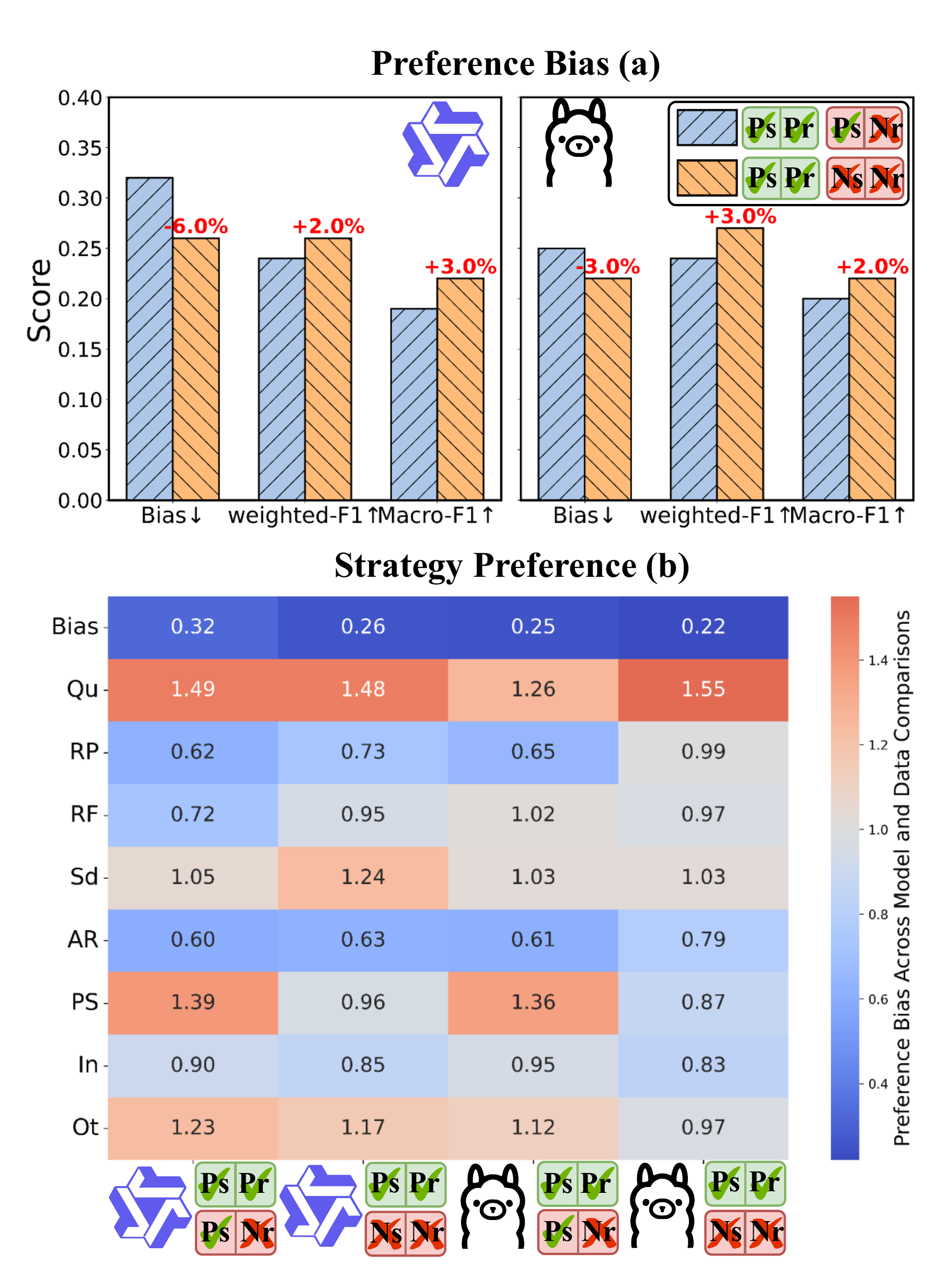}
    \caption{(a) Preference Bias and (b) Strategy Preference across Qwen and Llama models trained on different preference datasets.}
    \label{fig: Preliminary_3}
    
\end{figure}

\subsection{Limitations of the DPO Method (Obs 2)}
\label{sec: obs2}
To address \hyperref[sec: obs1]{Obs 1}, a natural approach is to treat filtered failures as negative signals and train with DPO. Prior work~\cite{zhao2025chain} adopted a vanilla DPO setup that jointly optimizes strategy-response pairs. However, as shown in Figure~\ref{fig: Introduction} and Figure~\ref{fig: Preliminary_2}, Vanilla-DPO relies heavily on the \textit{Question} strategy and shows a strong preference for it, which fails to significantly reduce preference bias (see section \hyperref[sec: Results]{Results}).

To investigate the failure of Vanilla-DPO in aligning with human preferences, we conduct a controlled study. We split the preference data into two types: \textbf{\texttt{\textcircled{1} (PsPr, PsNr)}}: where the preferred sample has both a positive strategy (Ps) and positive response (Pr), and the non-preferred sample has a positive strategy (Ps) but a negative response (Nr). \textbf{\texttt{\textcircled{2} (PsPr, NsNr)}}: where the non-preferred sample contains both negative strategy (Ns) and negative response (Nr).

We train models using each dataset on Qwen and Llama, and evaluate them on preference bias and strategy preference. As shown in Figure~\ref{fig: Preliminary_3} (a) and (b), models trained on \textbf{\texttt{\textcircled{2}}} consistently outperform those trained on \textbf{\texttt{\textcircled{1}}}. It reduces preference bias and better aligns with diverse strategies. 

\textbf{These results show that Vanilla-DPO training with entangled pairs like \textbf{\texttt{\textcircled{1}}} harms strategy learning.} This reveals two issues in Vanilla-DPO:

\begin{enumerate}
    \renewcommand{\labelenumi}{(\arabic{enumi})}
    \item \textbf{Entangled data structure:} The coupling between strategy and response complicates the construction of high-quality preference data, highlighting the need for more rigorous evaluation and filtering methods.
    
    \item \textbf{Optimization Ambiguity:} Entangled strategy and response training lead to optimization ambiguity or even negative optimization: mislabeling PsNr as a negative sample leads to negative optimization on strategy learning, while NsPr harms response learning.
\end{enumerate}

According to Hill’s Helping Skills Theory~\cite{hill1999helping} and Gross’s Extended Process Model of Emotion Regulation (EPMER)~\cite{gross2015emotion}, strategies should precede response generation and serve as its guidance. In essence, the two are decouplable. Inspired by this, we propose a decoupled modeling and staged optimization framework for ESC, which separates strategy planning from response generation, enabling more structured and targeted improvements in dialogue quality.


\begin{table}[t]

    \centering
    \scalebox{0.80}{
        \begin{tabular}{llccc}
            \toprule[1.2pt]
            & \textbf{Criteria} & \textbf{Total} & \textbf{Assistant} & \textbf{User} \\
            \midrule
            \multirow{4}{*}{\rotatebox{90}{\textbf{\texttt{ESConv}}}}
            & \# Dialogues & 1,040 & -- & -- \\
            & \# Utterances & 29,526 & 14,763 & 14,763 \\
            & Avg. Turns of Dialogue & 28.40 & 14.20 & 14.20 \\
            & Avg. Char of Utterance & 95.85 & 112.17 & 79.54 \\
            \bottomrule[1.2pt]
            & \textbf{Criteria} & \textbf{Total} & \textbf{Qwen} & \textbf{Llama} \\
            \midrule
            \multirow{7}{*}{\rotatebox{90}{\textbf{\texttt{IPM-PrefDial}}}}
            & \# Strategy Pref-Pairs & 21,370 & 10,651 & 10,719 \\
            & \# Response Pref-Pairs & 11,887 & 6,041 & 5,846 \\
            & Avg. Char of Chosen & 124.89 & 124.72 & 125.06 \\
            & Avg. Char of Rejected & 83.82 & 81.04 & 86.59 \\
            & \# Lack Emp. Response &  4,371 & 2,288 & 2,083 \\
            & \# Emo. Shift Response & 3,600 & 1,814 & 1,786 \\
            & \# Temp. Res. Response & 3,916 & 1,939 & 1,977 \\
            \bottomrule[1.2pt]
        \end{tabular}
    }
    \caption{Statistics of the \texttt{ESConv} and \texttt{IPM-PrefDial} Datasets. Char: Character, Pref-Pairs: Preference Pairs.}
    \label{tab: stat}
    
\end{table}

\section{Datasets}
\subsection{Preference Dataset Construction}
\paragraph{Expert-Guided ICL Annotation.}
\label{sec: icl}
Due to the large scale of the dataset, relying solely on human experts for psychological error annotation is highly time- and labor-intensive. Therefore, we adopt an In-Context Learning (ICL)~\cite{brown2020language} approach to guide LLM-based classification. Specifically, we engaged 3 professional psychologists to annotate representative examples of 5 common psychological errors, along with detailed explanations. These expert-labeled instances were then used as ICL prompts, enabling the LLM to perform classification in alignment with expert standards.

\paragraph{Inferential Preference Mining.}
The absence of failure-aware learning in standard SFT contributes to psychological errors observed in \hyperref[sec: obs1]{Obs 1}. To address these issues, we propose Inferential Preference Mining (IPM) for collecting high-quality preference data, and an Expert-Guided ICL approach that leverages LLMs' in-context learning to identify psychological errors.

Specifically, we first use the Qwen-SFT and Llama-SFT models to generate responses on the \texttt{ESConv} dataset. These responses are then paired with the corresponding gold responses to construct candidate preference pairs. To ensure consistency between the LLM and psychological expert annotators, we apply an Expert-Guided ICL Annotation process that prompts the LLM to filter out pairs exhibiting four common types of psychological errors. The remaining high-quality pairs constitute the \texttt{IPM-PrefDial} dataset. 

More concretely, for the Strategy Planner (SP), we pair suboptimal strategies $s_r$ exhibiting the psychological error \textit{(1) Strategy Mismatch} with gold strategies $s_c$ and context $c$ to form $D_{\text{SP-dpo}}$:
{
\begin{align}
D_{\text{SP-dpo}} = \left\{ \left( c^{(i)}, s_c^{(i)}, s_r^{(i)} \right) \right\}_{i=1}^{|D_{\text{SP-dpo}}|}.
\end{align}
}

For the Response Generator (RG), the SFT model produces multiple responses based on the gold strategy. Suboptimal responses $a_r$ exhibiting the psychological errors \textit{(2) Lack of Empathy}, \textit{(3) Early Emotion Shift}, or \textit{(4) Template Response} are identified via Expert-Guided ICL and paired with gold responses $a_c$ to form $D_{\text{RG-dpo}}$:
{
\begin{align}
D_{\text{RG-dpo}} = \left\{ \left( c^{(i)}, s^{(i)}, a_c^{(i)}, a_r^{(i)} \right) \right\}_{i=1}^{|D_{\text{RG-dpo}}|}.
\end{align}
}

\begin{figure*}[t]
    \centering
    \includegraphics[width=1\linewidth, trim=0 5 0 5, clip]{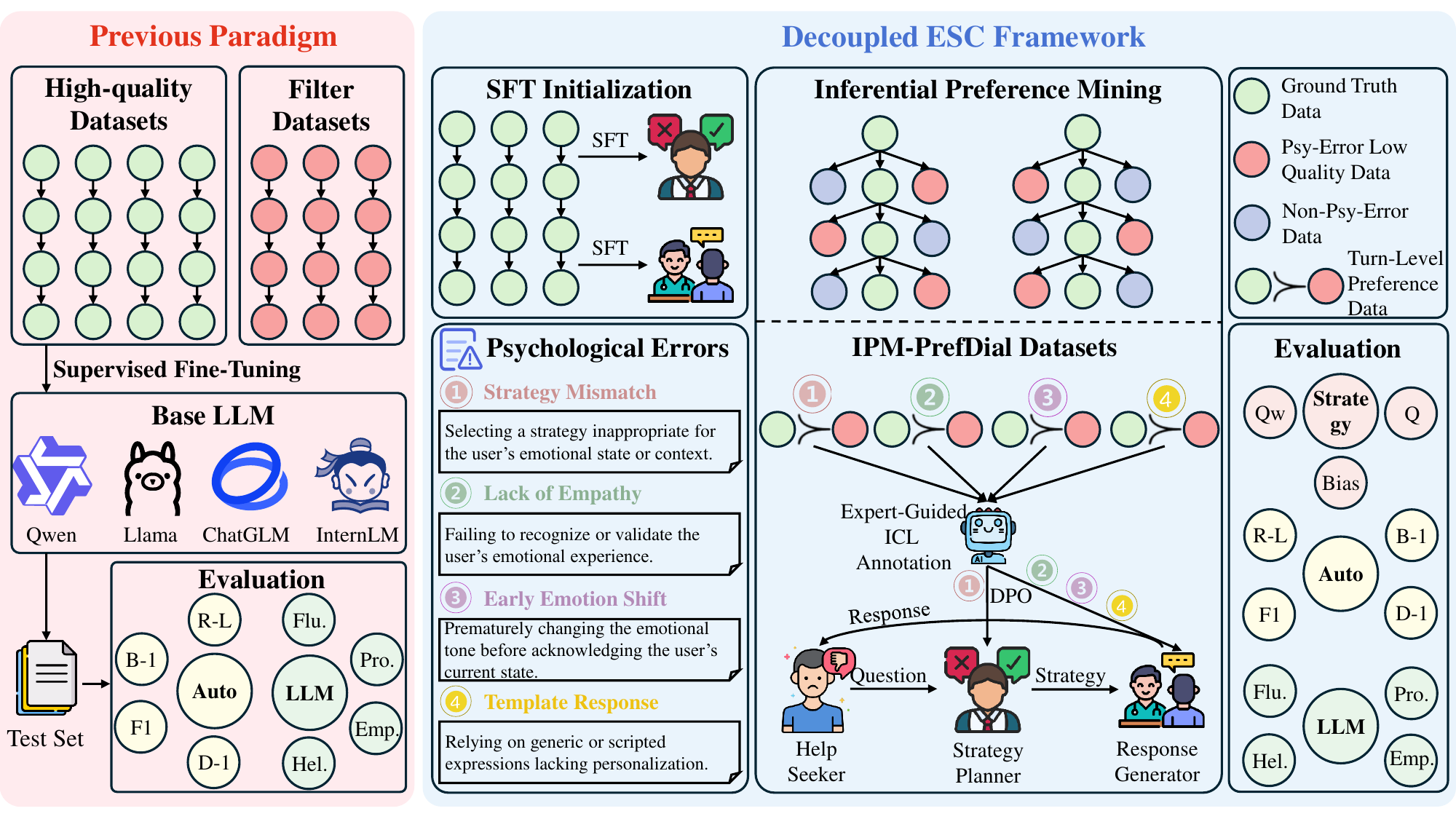}
    \caption{Comparison between previous vanilla SFT training paradigm and our proposed Decoupled ESC framework. The Decoupled ESC first undergoes SFT initialization, followed by DPO training using the \texttt{IPM-PrefDial} dataset.}
    \label{fig: Method}
    
\end{figure*}

\begin{table}[t]
    \centering
    \scalebox{0.75}{
    \begin{tabular}{lccccc}
        \toprule[1.2pt]
        \textbf{Model} & \textbf{Type} & \textbf{Flu.$\uparrow$} & \textbf{Pro.$\uparrow$} & \textbf{Emp.$\uparrow$} & \textbf{Hel.$\uparrow$} \\
        \midrule
        \multirow{3}{*}{\shortstack{Qwen\\2.5-7B-\\Instruct}}
        & Chosen& 3.82 & 3.52 & 3.20 & 2.95 \\
        & Rejected  & 3.65 & 3.09 & 2.41 & 2.32 \\
        &\cellcolor{yellow!20} \textit{Improve} ($\uparrow$) & \cellcolor{yellow!20}4.66\% & \cellcolor{yellow!20}13.92\% & \cellcolor{yellow!20}\textbf{32.78\%} & \cellcolor{yellow!20}27.16\% \\
        \midrule
        \multirow{3}{*}{\shortstack{Llama\\3.1-8B-\\Instruct}} 
        & Chosen& 3.99 & 3.74 & 3.33 & 3.09 \\
        & Rejected  & 3.93 & 3.21 & 2.40 & 2.39 \\
        & \cellcolor{yellow!20} \textit{Improve} ($\uparrow$) & \cellcolor{yellow!20}1.53\% & \cellcolor{yellow!20}16.52\% & \cellcolor{yellow!20}\textbf{38.75\%} & \cellcolor{yellow!20}29.29\% \\
        \bottomrule[1.2pt]
    \end{tabular}
    }
    \caption{LLM-based evaluation scores for chosen and rejected responses across four dimensions.}
    \label{tab: llm_eval}
\end{table}

\subsection{Datasets Statistics}
\paragraph{ESConv Dataset.}
We employ \texttt{ESConv} dataset for SFT training ($D_{\text{sft}}$), which includes $1,040$ dialogues with an average of $14.2$ turns and $95.9$ characters per turn. Strategy distribution and temporal trends are shown in Table~\ref{tab: support_strategy_by_model} and Figure~\ref{fig: data_statis_1} in the Appendix~\ref{sec: ESConv}.

\paragraph{IPM-PrefDial Dataset.}
\texttt{IPM-PrefDial} dataset contains $21,370$ strategy preference pairs ($D_{\text{SP-dpo}}$) and $11,887$ response preference pairs ($D_{\text{RG-dpo}}$). In $D_{\text{RG-dpo}}$, chosen responses average $124.89$ characters, rejected ones $83.82$. Major rejection reasons include \textit{Lack of Empathy} ($4,371$), \textit{Early Emotion Shift} ($3,600$), and \textit{Template Response} ($3,916$). Details are in Appendix~\ref{sec: IPM}.

\subsection{Datasets Quality}
We evaluate the content quality of $100$ samples from $D_{\text{RG-dpo}}$ using \texttt{gpt-4.1-mini-2025-04-14}\footnote{\url{https://openai.com/index/gpt-4-1}}. As shown in Table~\ref{tab: llm_eval}, chosen responses outperform rejected ones across four LLM-based metrics, with over a $30\%$ gain in \textit{Empathy}.

\begin{table*}[t]
\centering
\small
\renewcommand{\arraystretch}{1.2}
\setlength{\tabcolsep}{4pt}
\begin{tabular}{lcl|cccc|cccc|ccc}
\toprule[1.2pt]
\multirow{2}{*}{\textbf{Backbone}} & \multirow{2}{*}{\textbf{Paradigm}} & \multirow{2}{*}{\textbf{Method}} & \multicolumn{4}{c}{\textbf{Automatic Metrics.$\uparrow$}} & \multicolumn{4}{c}{\textbf{LLM-based Metrics.$\uparrow$}} & \multicolumn{3}{c}{\textbf{Strategy Metrics.}} \\
\cmidrule(lr){4-7} \cmidrule(lr){8-11} \cmidrule(lr){12-14}
& & & \textbf{D-1} & \textbf{B-1} & \textbf{F1} & \textbf{R-L} 
& \textbf{Flu.} & \textbf{Pro.} & \textbf{Emp.} & \textbf{Hel.} 
& \textbf{$\mathcal{B}\downarrow$} & \textbf{$\mathcal{Q_W}\uparrow$} & \textbf{$\mathcal{Q}\uparrow$} \\
\midrule
\multirow{9}{*}{\parbox{1.5cm}{\shortstack{Qwen2.5-\\7B-Instruct}}}
& \multirow{6}{*}{Vanilla} 
& \cellcolor{gray!3} Base & \cellcolor{gray!3}93.50 & \cellcolor{gray!3}9.75 & \cellcolor{gray!3}14.92 & \cellcolor{gray!3}12.59 & \cellcolor{gray!3}3.55 & \cellcolor{gray!3}2.53 & \cellcolor{gray!3}1.89 & \cellcolor{gray!3}1.38 & \cellcolor{gray!3}2.17 & \cellcolor{gray!3}8.41 & \cellcolor{gray!3}8.06 \\
& & \cellcolor{gray!5} +Direct-Refine & \cellcolor{gray!5}95.79 & \cellcolor{gray!5}10.91 & \cellcolor{gray!5}16.26 & \cellcolor{gray!5}14.35 & \cellcolor{gray!5}\textbf{4.17} & \cellcolor{gray!5}\textbf{2.97} & \cellcolor{gray!5}2.20 & \cellcolor{gray!5}1.77 & \cellcolor{gray!5}1.54 & \cellcolor{gray!5}13.52 & \cellcolor{gray!5}10.46 \\
& & \cellcolor{gray!5} +Self-Refine & \cellcolor{gray!5}97.04 & \cellcolor{gray!5}10.28 & \cellcolor{gray!5}15.85 & \cellcolor{gray!5}13.85 & \cellcolor{gray!5}3.68 & \cellcolor{gray!5}2.64 & \cellcolor{gray!5}1.94 & \cellcolor{gray!5}1.34 & \cellcolor{gray!5}1.45 & \cellcolor{gray!5}10.92 & \cellcolor{gray!5}9.63 \\
& & \cellcolor{gray!5} +Emotion CoT & \cellcolor{gray!5}\underline{97.17} & \cellcolor{gray!5}10.61 & \cellcolor{gray!5}16.07 & \cellcolor{gray!5}14.06 & \cellcolor{gray!5}3.95 & \cellcolor{gray!5}2.70 & \cellcolor{gray!5}\underline{2.50} & \cellcolor{gray!5}1.51 & \cellcolor{gray!5}1.87 & \cellcolor{gray!5}6.89 & \cellcolor{gray!5}6.63 \\
& & \cellcolor[HTML]{FFECEC} SFT & \cellcolor[HTML]{FFECEC}90.93 & \cellcolor[HTML]{FFECEC}15.61 & \cellcolor[HTML]{FFECEC}20.99 & \cellcolor[HTML]{FFECEC}17.78 & \cellcolor[HTML]{FFECEC}3.30 & \cellcolor[HTML]{FFECEC}2.61 & \cellcolor[HTML]{FFECEC}2.29 & \cellcolor[HTML]{FFECEC}\underline{2.12} & \cellcolor[HTML]{FFECEC}0.31 & \cellcolor[HTML]{FFECEC}24.89 & \cellcolor[HTML]{FFECEC}20.27 \\
& & \cellcolor[HTML]{FFD6D6} DPO & \cellcolor[HTML]{FFD6D6}88.13 & \cellcolor[HTML]{FFD6D6}16.23 & \cellcolor[HTML]{FFD6D6}21.24 & \cellcolor[HTML]{FFD6D6}18.03 & \cellcolor[HTML]{FFD6D6}3.47 & \cellcolor[HTML]{FFD6D6}2.67 & \cellcolor[HTML]{FFD6D6}2.36 & \cellcolor[HTML]{FFD6D6}\textbf{2.23} & \cellcolor[HTML]{FFD6D6}0.30 & \cellcolor[HTML]{FFD6D6}22.25 & \cellcolor[HTML]{FFD6D6}18.97 \\
\cmidrule(lr){2-14}
& \multirow{3}{*}{Decoupled} 
& \cellcolor{gray!3} Base & \cellcolor{gray!3}\textbf{97.55} & \cellcolor{gray!3}10.97 & \cellcolor{gray!3}16.33 & \cellcolor{gray!3}14.19 & \cellcolor{gray!3}3.92 & \cellcolor{gray!3}2.71 & \cellcolor{gray!3}2.17 & \cellcolor{gray!3}1.38 & \cellcolor{gray!3}1.92 & \cellcolor{gray!3}13.96 & \cellcolor{gray!3}12.07 \\
& & \cellcolor[HTML]{ECF4FF} SFT & \cellcolor[HTML]{ECF4FF}91.37 & \cellcolor[HTML]{ECF4FF}\underline{16.69} & \cellcolor[HTML]{ECF4FF}\underline{22.15} & \cellcolor[HTML]{ECF4FF}\underline{18.76} & \cellcolor[HTML]{ECF4FF}3.93 & \cellcolor[HTML]{ECF4FF}2.72 & \cellcolor[HTML]{ECF4FF}2.40 & \cellcolor[HTML]{ECF4FF}2.11 & \cellcolor[HTML]{ECF4FF}\underline{0.27} & \cellcolor[HTML]{ECF4FF}\underline{26.94} & \cellcolor[HTML]{ECF4FF}\underline{21.37} \\
& & \cellcolor[HTML]{DAE8FC} DPO & \cellcolor[HTML]{DAE8FC}89.84 & \cellcolor[HTML]{DAE8FC}\textbf{17.73} & \cellcolor[HTML]{DAE8FC}\textbf{22.86} & \cellcolor[HTML]{DAE8FC}\textbf{19.31} & \cellcolor[HTML]{DAE8FC}\underline{3.99} & \cellcolor[HTML]{DAE8FC}\underline{2.90} & \cellcolor[HTML]{DAE8FC}\textbf{2.54} & \cellcolor[HTML]{DAE8FC}2.02 & \cellcolor[HTML]{DAE8FC}\textbf{0.22} & \cellcolor[HTML]{DAE8FC}\textbf{27.09} & \cellcolor[HTML]{DAE8FC}\textbf{21.77} \\
\midrule
\multirow{9}{*}{\parbox{1.5cm}{\shortstack{Llama3.1-\\8B-Instruct}}}
& \multirow{6}{*}{Vanilla} 
& \cellcolor{gray!3} Base & \cellcolor{gray!3}\textbf{95.09} & \cellcolor{gray!3}12.38 & \cellcolor{gray!3}16.85 & \cellcolor{gray!3}14.01 & \cellcolor{gray!3}\textbf{4.35} & \cellcolor{gray!3}\underline{3.21} & \cellcolor{gray!3}2.36 & \cellcolor{gray!3}1.76 & \cellcolor{gray!3}1.03 & \cellcolor{gray!3}15.74 & \cellcolor{gray!3}14.09  \\
& & \cellcolor{gray!5} +Direct-Refine & \cellcolor{gray!5}90.07 & \cellcolor{gray!5}11.36 & \cellcolor{gray!5}14.97 & \cellcolor{gray!5}12.79 & \cellcolor{gray!5}3.35 & \cellcolor{gray!5}2.82 & \cellcolor{gray!5}2.16 & \cellcolor{gray!5}1.35 & \cellcolor{gray!5}1.72 & \cellcolor{gray!5}12.12 & \cellcolor{gray!5}9.98 \\
& & \cellcolor{gray!5} +Self-Refine & \cellcolor{gray!5}87.18 & \cellcolor{gray!5}10.72 & \cellcolor{gray!5}14.26 & \cellcolor{gray!5}12.20 & \cellcolor{gray!5}3.53 & \cellcolor{gray!5}2.95 & \cellcolor{gray!5}2.40 & \cellcolor{gray!5}1.45 & \cellcolor{gray!5}1.68 & \cellcolor{gray!5}13.93 & \cellcolor{gray!5}12.00 \\
&& \cellcolor{gray!5} +Emotion CoT & \cellcolor{gray!5}77.32 & \cellcolor{gray!5}10.06 & \cellcolor{gray!5}13.32 & \cellcolor{gray!5}11.33 & \cellcolor{gray!5}3.24 & \cellcolor{gray!5}2.88 & \cellcolor{gray!5}\underline{2.56} & \cellcolor{gray!5}1.63 & \cellcolor{gray!5}1.86 & \cellcolor{gray!5}13.31 & \cellcolor{gray!5}11.35 \\
& & \cellcolor[HTML]{FFECEC} SFT & \cellcolor[HTML]{FFECEC}91.29 & \cellcolor[HTML]{FFECEC}15.75 & \cellcolor[HTML]{FFECEC}21.38 & \cellcolor[HTML]{FFECEC}18.11 & \cellcolor[HTML]{FFECEC}3.31 & \cellcolor[HTML]{FFECEC}2.52 & \cellcolor[HTML]{FFECEC}2.22 & \cellcolor[HTML]{FFECEC}2.06 & \cellcolor[HTML]{FFECEC}0.26 & \cellcolor[HTML]{FFECEC}24.54 & \cellcolor[HTML]{FFECEC}19.97 \\
& & \cellcolor[HTML]{FFD6D6} DPO & \cellcolor[HTML]{FFD6D6}91.25 & \cellcolor[HTML]{FFD6D6}15.15 & \cellcolor[HTML]{FFD6D6}20.49 & \cellcolor[HTML]{FFD6D6}17.25 & \cellcolor[HTML]{FFD6D6}3.41 & \cellcolor[HTML]{FFD6D6}2.79 & \cellcolor[HTML]{FFD6D6}2.41 & \cellcolor[HTML]{FFD6D6}\textbf{2.28} & \cellcolor[HTML]{FFD6D6}0.28 & \cellcolor[HTML]{FFD6D6}24.00 & \cellcolor[HTML]{FFD6D6}19.89 \\
\cmidrule(lr){2-14}
& \multirow{3}{*}{Decoupled} 
& \cellcolor{gray!3} Base & \cellcolor{gray!3}\underline{94.65} & \cellcolor{gray!3}12.67 & \cellcolor{gray!3}16.70 & \cellcolor{gray!3}14.01 & \cellcolor{gray!3}\underline{4.24} & \cellcolor{gray!3}\textbf{3.24} & \cellcolor{gray!3}2.34 & \cellcolor{gray!3}1.66 & \cellcolor{gray!3}1.62 & \cellcolor{gray!3}7.54 & \cellcolor{gray!3}7.67 \\
& & \cellcolor[HTML]{ECF4FF} SFT & \cellcolor[HTML]{ECF4FF}91.51 & \cellcolor[HTML]{ECF4FF}\underline{16.97} & \cellcolor[HTML]{ECF4FF}\underline{22.42} & \cellcolor[HTML]{ECF4FF}\underline{19.12} & \cellcolor[HTML]{ECF4FF}3.87 & \cellcolor[HTML]{ECF4FF}2.74 & \cellcolor[HTML]{ECF4FF}2.39 & \cellcolor[HTML]{ECF4FF}1.95 & \cellcolor[HTML]{ECF4FF}\underline{0.23} & \cellcolor[HTML]{ECF4FF}\underline{26.03} & \cellcolor[HTML]{ECF4FF}\underline{21.36} \\
& & \cellcolor[HTML]{DAE8FC} DPO & \cellcolor[HTML]{DAE8FC}90.35 & \cellcolor[HTML]{DAE8FC}\textbf{17.50} & \cellcolor[HTML]{DAE8FC}\textbf{22.59} & \cellcolor[HTML]{DAE8FC}\textbf{19.16} & \cellcolor[HTML]{DAE8FC}3.81 & \cellcolor[HTML]{DAE8FC}2.73 & \cellcolor[HTML]{DAE8FC}\textbf{2.64} & \cellcolor[HTML]{DAE8FC}\underline{2.17} & \cellcolor[HTML]{DAE8FC}\textbf{0.15} & \cellcolor[HTML]{DAE8FC}\textbf{27.10} & \cellcolor[HTML]{DAE8FC}\textbf{22.94} \\
\bottomrule[1.2pt]
\end{tabular}
\caption{Comparison of models under different optimization paradigms and training methods. The best score is \textbf{in-bold}, while the second best score is \underline{underlined}. $\uparrow$ means a higher score is better whereas $\downarrow$ is exactly the opposite.}
\label{tab: results}
\end{table*}

\section{Methodology}
\subsection{Decoupled ESC Framework}
To address the issue raised in \hyperref[sec: obs1]{Obs 1}  and \hyperref[sec: obs2]{Obs 2}, that vanilla training of strategy planning and response generation can lead to negative optimization, hindering the reduction of preference bias and the improvement of response quality. As shown in Figure~\ref{fig: Method}, we propose a \textbf{Decoupled ESC optimization framework}, inspired by the Extended Process Model of Emotion Regulation (EPMER)~\cite{gross2015emotion}, which divides emotion regulation into three sequential stages: \textit{Identification}, \textit{Strategy Selection}, and \textit{Implementation}. We decouple the ESC generation process into two independent subtasks: Strategy Planning and Response Generation. This enables more stable and controllable training for each.

Specifically, we adopt a decoupled two-stage modeling framework: a Strategy Planner selects a optimal strategy based on the dialog history \( c_t = (u_0, a_0, \ldots, u_{t-1}, a_{t-1}, u_t) \), where \( u \) and \( a \) denote user and assistant utterances, respectively. The strategy is generated as \( s_t \sim \text{LLM}_{\text{SP}}(s \mid c_t) \). Then, a Response Generator generates an empathic reply conditioned on both the selected strategy and the dialog context: \( a_t \sim \text{LLM}_{\text{RG}}(a \mid c_t, s_t) \).

\subsection{Decoupled-SFT and Decoupled-DPO}

\paragraph{Decoupled-SFT.}
To optimize the performance of the Strategy Planner and Response Generator, we first initialize these two modules using the SFT method to endow them with the capabilities for strategy planning and empathic response generation. Specifically, based on real dialogues from the ESC dataset, we constructed a turn-level training dataset $D_{\text{sft}} = \left\{ \left( c^{(i)}, s^{(i)}, a^{(i)} \right) \right\}_{i=1}^{|D_{\text{sft}}|}$. The two modules are then fine-tuned separately using SFT: (1) Strategy Planner: Using the dialogue context \( c \) and the supporter’s response strategy \( s \), we perform turn-level training to minimize the loss function:
{
\begin{align}
\mathcal{L}_{\text{SP-sft}} = - \mathbb{E}_{(c, s) \sim D_{\text{sft}}} \left[ \log \mathrm{LLM}_{\text{SP}}(s | c) \right].
\end{align}
}(2) Response Generator: Given the context $c$, strategy $s$, and response $a$, minimizing the loss:
{
\begin{align}
\mathcal{L}_{\text{RG-sft}} = - \mathbb{E}_{(c, s, a) \sim D_{\text{sft}}} \left[ \log \mathrm{LLM}_{\text{RG}}(a | c, s) \right].
\end{align}
}

\paragraph{Decoupled-DPO.}
To reduce psychological errors, we further optimize the Strategy Planner and Response Generator using the offline RL method (DPO). Based on the preference dataset \texttt{IPM-PrefDial}, which includes $D_{\text{SP-dpo}}$ and $D_{\text{RG-dpo}}$, we separately train both modules to enhance strategy selection and response generation. For the Strategy Planner, we apply DPO on $D_{\text{SP-dpo}}$ to encourage preference for gold strategies and reduce bias toward suboptimal ones. The loss function is defined as:

{\small
\begin{align}
\mathcal{L}_{\text{SP-dpo}} = - \mathbb{E}_{(c, s_c, s_r) \sim D_{\text{SP-dpo}}} \left[
\log \sigma \Bigg(
\beta \log \frac{\pi_\theta(s_c|c)}{\pi_{\text{ref}}(s_c|c)} \right. \nonumber \\
\left. - \beta \log \frac{\pi_\theta(s_r|c)}{\pi_{\text{ref}}(s_r|c)}
\Bigg) \right],
\end{align}
}

For the Response Generator, we apply DPO on $D_{\text{RG-dpo}}$ to improve response quality and empathy, with the loss function defined as:

{\small
\begin{align}
    \mathcal{L}_{\text{RG-dpo}} = - \mathbb{E}_{(c, s, a_c, a_r) \sim D_{\text{RG-dpo}}} \left[
    \log \sigma \left(
    \beta \log \frac{\pi_\theta(a_c|c,s)}{\pi_{\text{ref}}(a_c|c,s)} \right. \right. \notag \\
    \left. \left. 
    - \beta \log \frac{\pi_\theta(a_r|c,s)}{\pi_{\text{ref}}(a_r|c,s)}
    \right) \right].
\end{align}
}

where $\pi_\theta$ denotes the model being optimized, and $\pi_{\text{ref}}$ denotes the reference model after SFT.


\section{Experiments}
In this section, we conduct extensive experiments to address the following research questions:
\begin{itemize}[itemsep=0.01em, leftmargin=8pt]
    \item \textbf{RQ1:} What are the performance differences between DPO and SFT in bias mitigation and response generation within a coupled framework?
    \item \textbf{RQ2:} What specific advantages does decoupling strategy planning and response generation bring to ESC tasks?
    \item \textbf{RQ3:} In the Decoupled ESC framework, how do SFT and DPO respectively affect the model's bias and generation quality?
    \item \textbf{RQ4:} To what extent can the Decoupled ESC framework effectively reduce psychological errors?
\end{itemize}

\subsection{Experimental Setup}
\paragraph{Backbones.} 
We conducted experiments using two LLMs: \texttt{Qwen2.5-7B-Instruct}~\cite{qwen2.5} and \texttt{Llama3.1-8B-Instruct}~\cite{dubey2024llama}.

\paragraph{Baselines.} 
We compare vanilla coupled models (Base, SFT, DPO) with prompt-optimization baselines such as Direct-Refine, Self-Refine~\cite{madaan2023self}, and Emotional CoT~\cite{wei2022chain}.

\paragraph{Datasets.} 
The \texttt{ESConv}~\cite{liu2021towards} dataset is split into train, valid, and test sets in an 8:1:1 ratio, with the training set used for SFT. The \texttt{IPM-PrefDial} dataset is used for DPO training.

\paragraph{Evaluation Metrics.}
We evaluate model performance using the following metrics: 
\textbf{(1) Automatic Metrics}, including BLEU-1 (B-1)~\cite{papineni2002bleu}, Distinct-1 (D-1)~\cite{li2015diversity}, F1-score (F1), and ROUGE-L (R-L)~\cite{lin2004rouge}; 
\textbf{(2) LLM-based Metrics}, including \textit{Fluency (Flu.)}, \textit{Professionalism (Pro.)}, \textit{Empathy (Emp.)}, and \textit{Helpfulness (Hel.)}. All metrics are rated on a 5-point Likert scale~\cite{joshi2015likert}; 
\textbf{(3) Strategy Metrics}, including preference bias ($\mathcal{B}$)~\cite{kang2024can} and strategy prediction accuracy (weighted-F1 $Qw$ and Macro-F1 $Q$). Detailed definitions of the evaluation metrics, the prompt, and the Bias calculation formula are provided in Appendix~\ref{sec: EvaDetail}.

\paragraph{Implementation Details.}
For SFT, we train for 3 epochs using a batch size of 32 and a learning rate of 1e-5. DPO is trained for 1 epoch under the same settings. All experiments are conducted on 4×24GB RTX 4090 GPUs. For LLM-based evaluation, we randomly sample 100 test instances and evaluate them using \texttt{gpt-4.1-mini-2025-04-14}\footnotemark[4] alongside 3 psychology experts. Additional implementation details are provided in Appendix~\ref{sec: Experiment}.

\begin{table}[t]
\centering
\renewcommand{\arraystretch}{1}
\setlength{\tabcolsep}{2pt}
\scalebox{0.78}{
\begin{tabular}{l|cc|cccc|c}
\toprule[1.2pt]
\textbf{Model} 
& \textbf{B-1$\uparrow$} & \textbf{R-L$\uparrow$} 
& \textbf{Flu.$\uparrow$} & \textbf{Pro.$\uparrow$} & \textbf{Emp.$\uparrow$} & \textbf{Hel.$\uparrow$} 
& \textbf{$\mathcal{B}\downarrow$} \\
\midrule
Vanilla-SFT           & 15.75 & 18.11 & 3.31 & 2.52 & 2.22 & \textbf{2.06} & 0.26 \\
Decoupled-SFT         & \textbf{16.97} & \textbf{19.12} & \textbf{3.87} & \textbf{2.74} & \textbf{2.39} & 1.95 & \textbf{0.23} \\

\midrule
Vanilla-DPO           & 15.15 & 17.25 & 3.41 & \textbf{2.79} & 2.41 & \textbf{2.28} & 0.28 \\
Decoupled-DPO & \textbf{17.50} & \textbf{19.16} & \textbf{3.81} & 2.73 & \textbf{2.64} & 2.17 & \textbf{0.15} \\
\bottomrule[1.2pt]
\end{tabular}
}
\caption{Results of SFT and DPO under Vanilla and Decoupled Paradigms for \texttt{Llama3.1-8B-Instruct}.}
\label{tab: improvement}
\end{table}

\subsection{Experimental Results}
\label{sec: Results}

\paragraph{Vanilla-DPO vs. Vanilla-SFT (RQ1).} 
As shown in Table~\ref{tab: results}, under the vanilla setting, DPO consistently outperforms SFT on LLM-based metrics for both Qwen and Llama, indicating enhanced response quality. 

However, Strategy-Metrics evaluation shows mixed results: DPO slightly reduces bias ($\mathcal{B}$) on Qwen but increases bias and lowers accuracy on Llama. We attribute this to DPO’s sensitivity to noisy data in the vanilla setting—conflicting optimization signals between strategy and content can cause negative transfer (see section \hyperref[sec: obs2]{Obs 2}). We further compare the impact of different preference data on coupled models in Appendix~\ref{sec: Analysis}.

\paragraph{Decoupled vs. Vanilla (RQ2).} 
As shown in Tables~\ref{tab: results} and~\ref{tab: improvement}, decoupled models outperform vanilla ones on most metrics. Notably, the preference bias of Decoupled-DPO models drops to $0.22$ (Qwen) and $0.15$ (Llama), much better than Vanilla-DPO and Vanilla-SFT. This shows decoupled optimization effectively reduces preference bias.

\begin{figure}[t]
    \centering
    \includegraphics[width=0.90\linewidth, trim=0 80 0 0, clip]{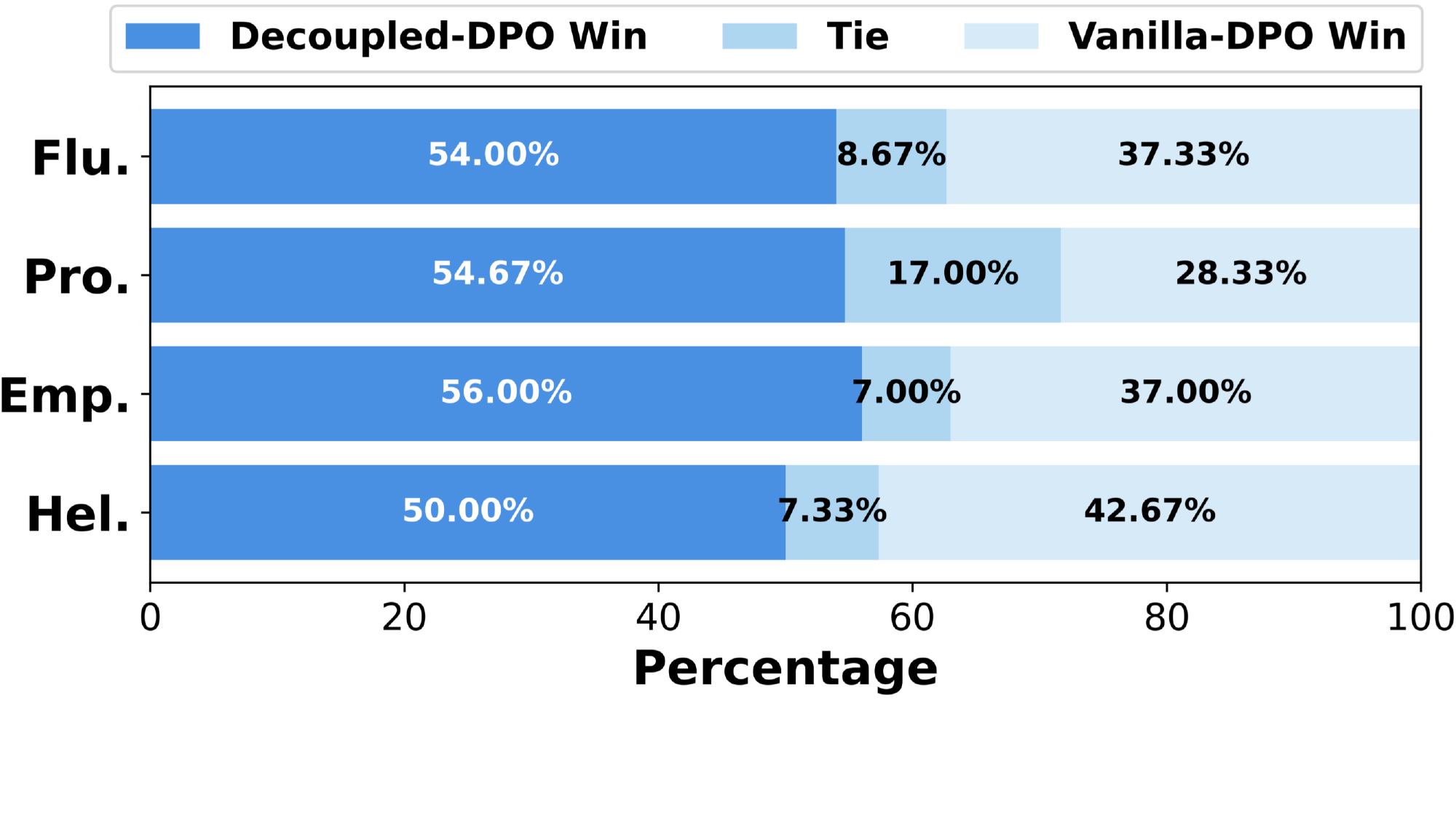}
    \caption{Comparison of Human-Evaluated Win Rates for Decoupled-DPO (Llama) and Vanilla-DPO (Llama).}
    \label{fig: winloss}
\end{figure}

\paragraph{\textit{Human Evaluation}}
We recruited 3 licensed psychology experts to independently evaluate 100 dialogues from Decoupled-DPO (Llama). The inter-rater agreement, measured by Fleiss’ Kappa ($\kappa$)~\cite{fleiss1973equivalence}, indicates moderate consistency across 4 LLM-based Metrics: \textit{Flu.} ($0.414$), \textit{Pro.} ($0.398$), \textit{Emp.} ($0.421$), and \textit{Hel.} ($0.368$). Additionally, we also performed a pairwise win/loss comparison between Decoupled-DPO and Vanilla-DPO (both Llama) on 100 samples, with evaluator agreement again measured by $\kappa$. As shown in Figure~\ref{fig: winloss}, Decoupled-DPO consistently outperforms Vanilla-DPO across all 4 LLM-based Metrics, with $\kappa$ scores of $0.617$ (\textit{Flu.}), $0.470$ (\textit{Pro.}), $0.450$ (\textit{Emp.}), and $0.431$ (\textit{Hel.}), indicating substantial agreement.

We attribute this to the decoupled framework. Its strategy planning module, as shown by Kang~\textit{et al.}~\cite{kang2024can}, acts as an external planner that helps reduce strategy preference bias. It also avoids the optimization conflict in vanilla training and simplifies preference data construction.

\paragraph{Decoupled-DPO vs. Decoupled-SFT (RQ3).} 
As shown in Tables~\ref{tab: results} and~\ref{tab: improvement}, DPO brings greater improvements over SFT in the decoupled setting, with bias reduced from $0.27$ to $0.22$ for Qwen, and from $0.23$ to $0.15$ for Llama. In contrast, the vanilla framework yields minimal bias reduction from SFT to DPO, highlighting the stronger synergy between decoupling and DPO training. Table~\ref{tab: ablation study} further shows that, given ground-truth strategies, the DPO-trained Response Generator consistently outperforms its SFT counterpart across all LLM-based metrics.
\begin{table}[t]
    \centering
    \setlength{\tabcolsep}{3pt}
    \scalebox{0.8}{
    \begin{tabular}{lcccccccc}
        \toprule[1.2pt]
        \textbf{Backbone} & \textbf{GT} & \textbf{SFT} & \textbf{DPO} & \textbf{Flu.$\uparrow$} & \textbf{Pro.$\uparrow$} & \textbf{Emp.$\uparrow$} & \textbf{Hel.$\uparrow$} \\ \midrule
        \multirow{2}{*}{\shortstack{Qwen2.5-\\7B-Instruct}} & \ding{51} & \ding{51} & \ding{55} & 3.66  & 3.02  & 2.51  & 2.37  \\
                               & \ding{51} & \ding{51} & \ding{51} & \textbf{3.77}  & \textbf{3.26}  & \textbf{2.75}  & \textbf{2.67}  \\ \midrule
        \multirow{2}{*}{\shortstack{Llama3.1-\\8B-Instruct}} & \ding{51} & \ding{51} & \ding{55} & 3.67  & 3.18  & 2.71  & 2.52  \\
                                & \ding{51} & \ding{51} & \ding{51} & \textbf{3.94}  & \textbf{3.44}  & \textbf{2.90}  & \textbf{2.73}  \\ \bottomrule[1.2pt]
    \end{tabular}
    }
    \caption{Ablation study on Response Generator. \ding{51} indicates that the method or data is used in training, while \ding{55} indicates it is not. GT: Ground Truth Strategy.}
    \label{tab: ablation study}
    
\end{table}

\begin{figure}[h]
    \centering
    \includegraphics[width=1\linewidth, trim=30 8 30 8, clip]{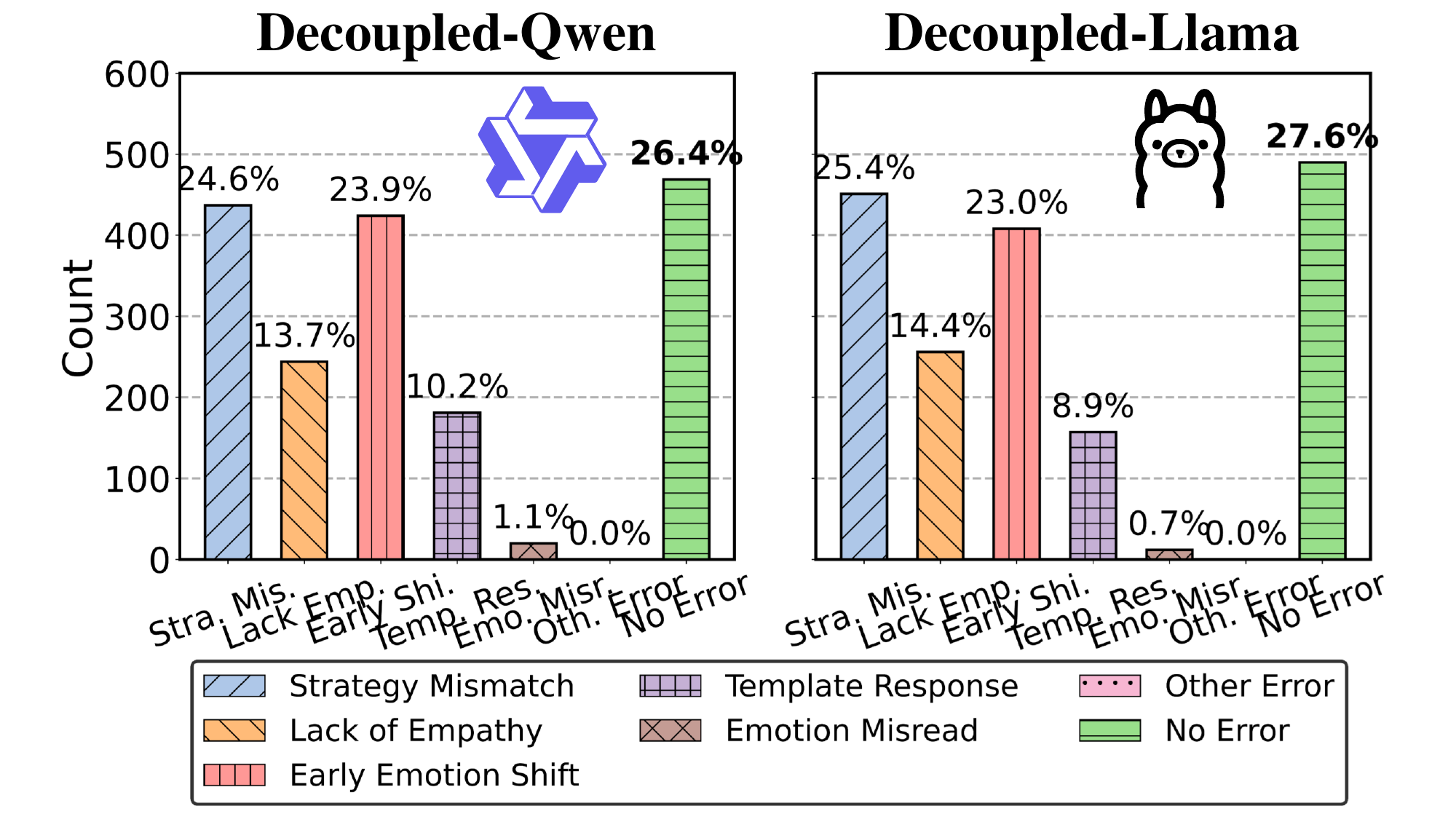}
    \caption{Proportion of psy-error types in the response of Qwen and Llama under the decoupled framework.}
    \label{fig: RQ4}
    
\end{figure}

\paragraph{Decoupled-DPO on Psychological Errors (RQ4).} 
To verify the effectiveness of Decoupled-DPO in improving response quality, we adopt the same error categorization approach as \hyperref[sec: obs1]{Obs 1}. By contrasting Figures~\ref{fig: Preliminary} and~\ref{fig: RQ4}, we observe that Decoupled-DPO achieves the highest proportion of \textit{No Error} cases at $27$\%, outperforming Qwen-SFT and Llama-SFT by an average of $7$\%. This demonstrates that Decoupled-DPO effectively reduces common psychological errors and improves overall response quality.


\section{Conclusion}
In this paper, we propose a Decoupled ESC framework that separates strategy planning from empathetic response generation, enabling targeted optimization and avoiding mutual interference. Extensive experiments demonstrate that our Decoupled ESC framework significantly outperforms joint optimization baselines, effectively reducing preference bias and improving response quality in Emotional Support Conversation tasks.

\section*{Limitations}
Our study, while demonstrating promising results, has several limitations that suggest avenues for future research. The annotation of psychological errors relied on an in-context learning approach to manage the resource-intensive nature of expert labeling; while efficient, this method's classifications may not be perfectly consistent with expert judgment. Furthermore, our experiments were constrained by computational resources to models up to 9B parameters, so the efficacy of our method on much larger models (e.g., 70B) requires future validation. The generalizability of our decoupled framework also warrants broader investigation, as it was primarily tested with DPO and should be assessed with other algorithms like KTO, SimPO, and IPO. Finally, this work is confined to textual analysis, and a key future direction is to extend our framework to incorporate multimodal inputs such as speech and facial expressions for more holistic emotional support.

\section*{Ethics Statement}
\subsection*{Data Usage Agreement}
This research utilizes the \texttt{ESConv} and \texttt{FailedESConv} dataset~\cite{liu2021towards}, which has been obtained with proper authorization and in compliance with data usage agreements. We ensure that all data used in this study is handled responsibly and in accordance with ethical standards, respecting the privacy and confidentiality of individuals involved. All necessary agreements and permissions for the use of this dataset have been signed, ensuring full compliance with data protection regulations.

\subsection*{Model Usage Policy}
It should be noted that while the model demonstrates certain capabilities in psychological support tasks, its strategies cannot encompass the full range of approaches and techniques used in real-life professional counseling. Given the diversity of users’ emotional states and circumstances, the model's responses may not always align with professional standards and may unintentionally affect users’ emotional well-being. Therefore, this model is intended for academic research only and is not recommended for commercial use. Caution is advised when using it beyond research settings, and it should not be applied to real-world counseling without professional supervision.

\section*{Acknowledgements}
This work was supported by the National Natural Science Foundation of China (62293554, U2336212), “Pioneer” and “Leading Goose” R\&D Program of Zhejiang (2024C01073), Ningbo Innovation “Yongjiang 2035” Key Research and Development Programme (2024Z292), and Young Elite Scientists Sponsorship Program by CAST (2023QNRC001).

\bibliography{custom}
\bibliographystyle{acl_natbib}

\clearpage

\appendix
\section{Definitions}
\label{sec: Definition}
\subsection{Definitions of Gross’s Extended Process Model of Emotion Regulation}
\label{sec: Definition_Gross}
The Extended Process Model of Emotion Regulation (EPMER), proposed by Gross in 2015~\cite{gross2015emotion}, refines earlier models by conceptualizing emotion regulation as a temporally ordered process comprising 3 core stages:
\begin{enumerate}
    \item \textbf{Identification:} Individuals assess whether an emotional response needs to be regulated based on situational goals and personal relevance.
    \item \textbf{Selection:} A regulation strategy is chosen from available options, guided by the expected outcome of regulating the emotion.
    \item \textbf{Implementation:} The selected strategy is carried out and monitored.
\end{enumerate}

\subsection{Definitions of Psychological Errors}
\label{sec: PsyError}
Under the guidance of 3 psychological experts and based on psychological literature~\cite{raskin2005person, stebnicki2007empathy}, we identified common errors frequently made by psychological experts in real-world therapy sessions. These were categorized into 5 types of empathy-related psychological errors\footnote{All definitions of psychological errors were reviewed by 3 psychological experts.}:
\begin{itemize}
    \item \textbf{Strategy Mismatch}: Selecting a strategy inappropriate for the user’s emotional state or context.
    \item \textbf{Lack of Empathy}: Failing to recognize or validate the user’s emotional experience.
    \item \textbf{Early Emotion Shift}: Prematurely changing the emotional tone before acknowledging the user’s current state.
    \item \textbf{Template Response}: Relying on generic or scripted expressions lacking personalization.
    \item \textbf{Emotion Misread}: Misinterpreting the user’s emotional cues, leading to unaligned responses.
\end{itemize}
Figures~\ref{fig: Strategy_Mismatch}, \ref{fig: Lack_Empathy}, \ref{fig: EarlyEmotionShift}, and \ref{fig: TemplateResponse} illustrate representative examples of the first four error types, drawn from rejected responses in the \texttt{IPM-PrefDial} dataset.

\subsection{Definitions of Counseling Stages}
\label{sec: Definition_Stage}
Liu~\textit{et al.}~\cite{liu2021towards} developed a three-stage counseling framework based on Hill's Helping Skills Theory~\cite{hill1999helping}.

\begin{enumerate}
    \item \textbf{Exploration}: Explore to identify the help-seeker's problem.
    \item \textbf{Comforting}: Comfort the help-seeker by expressing empathy and understanding.
    \item \textbf{Action}: Assist the help-seeker in solving their problems.
\end{enumerate}

Although most cases in our dataset follow the counseling sequence of (1) Exploration → (2) Comforting → (3) Action, some cases are adjusted based on the help-seeker's specific situation.

\begin{table*}[t]
\centering
\resizebox{1\linewidth}{!}{
\renewcommand{\arraystretch}{1.2}
\setlength{\tabcolsep}{4pt}
    \begin{tabular}{l|c|ccc|cccc|cccc|ccc}
        \toprule[1.2pt]
        \multirow{2}{*}{\textbf{Backbone}} & \multicolumn{1}{c}{\textbf{Chosen}} & \multicolumn{3}{c}{\textbf{Rejected}} & \multicolumn{4}{c}{\textbf{Automatic Metrics.$\uparrow$}} & \multicolumn{4}{c}{\textbf{LLM-based Metrics.$\uparrow$}} & \multicolumn{3}{c}{\textbf{Strategy Metrics.}} \\
        \cmidrule(lr){2-2} \cmidrule(lr){3-5} \cmidrule(lr){6-9} \cmidrule(lr){10-13} \cmidrule(lr){14-16}
        & \textbf{PsPr} & \textbf{NsNr} & \textbf{PsNr} & \textbf{NsPr} & \textbf{D-1} & \textbf{B-1} & \textbf{F1} & \textbf{R-L} & \textbf{Flu.} & \textbf{Pro.} & \textbf{Emp.} & \textbf{Hel.} & \textbf{$\mathcal{B}\downarrow$} & \textbf{$\mathcal{Q_W}\uparrow$} & \textbf{$\mathcal{Q}\uparrow$} \\
        \midrule
        \multirow{4}{*}{\parbox{1.5cm}{\shortstack{Qwen2.5-\\7B-Instruct}}}
        & \ding{51} & \ding{51} & \ding{55} & \ding{55} & 89.38 & \underline{15.93} & \underline{20.94} & \underline{17.75} & 3.31 & 2.64 & 2.34 & 2.15 & \textbf{0.26} & \textbf{25.60} & \textbf{21.69} \\
        & \ding{51} & \ding{51} & \ding{51} & \ding{55} & \underline{89.45} & 15.73 & 20.80 & 17.50 & \textbf{3.50} & \textbf{2.73} & \textbf{2.53} & \textbf{2.25} & 0.30 & 22.81 & 18.76 \\
        & \ding{51} & \ding{51} & \ding{55} & \ding{51} & \textbf{90.83} & 15.32 & 20.78 & 17.53 & 3.19 & 2.54 & 2.17 & 2.13 & \underline{0.29} & \underline{22.91} & 18.63  \\
        & \ding{51} & \ding{51} & \ding{51} & \ding{51} & 88.13 & \textbf{16.23} & \textbf{21.24} & \textbf{18.03} & \underline{3.47} & \underline{2.67} & \underline{2.36} & \underline{2.23} & 0.30 & 22.25 & \underline{18.97} \\
        \midrule
        \multirow{4}{*}{\parbox{1.5cm}{\shortstack{Llama3.1-\\8B-Instruct}}}
        & \ding{51} & \ding{51} & \ding{55} & \ding{55} & 90.72 & \textbf{16.08} & \underline{21.41} & \textbf{18.07} & 3.45 & \underline{2.69} & 2.38 & 2.22 & \textbf{0.22} & \textbf{26.69} & \textbf{21.76} \\
        & \ding{51} & \ding{51} & \ding{51} & \ding{55} & \underline{91.19} & \underline{15.92} & 21.30 & 17.92 & \underline{3.48} & 2.68 & \textbf{2.45} & \underline{2.26} & 0.29 & 23.82 & 19.71\\
        & \ding{51} & \ding{51} & \ding{55} & \ding{51} & \underline{91.19} & \underline{15.92} & \textbf{21.45} & \underline{18.01} & \textbf{3.49} & 2.65 & 2.22 & 2.15 & \textbf{0.22} & \underline{25.20} & \underline{21.07} \\
        & \ding{51} & \ding{51} & \ding{51} & \ding{51} & \textbf{91.25} & 15.15 & 20.49 & 17.25 & 3.41 & \textbf{2.79} & \underline{2.41} & \textbf{2.28} & \underline{0.28} & 24.00 & 19.89 \\
        \bottomrule[1.2pt]
    \end{tabular}}
\caption{Comparison of coupled models trained with different preference data.  \ding{51} denotes that the training set contains this type of data, while \ding{55} denotes its absence in the training set. The best score is \textbf{in-bold}, while the second best score is \underline{underlined}.}
\label{tab: Comparison of coupled models}

\end{table*}

\subsection{Definitions of Strategies}
\label{sec: Definition_Strategies}
The strategies and its definitions in this study align with Liu \textit{et al.}~\cite{liu2021towards} and follow Hill's Helping Skills Theory~\cite{hill1999helping}.

\begin{itemize}
    \item \textbf{Question (Qu)}: Asking for information related to the problem to help the help-seeker articulate the issues that they face. Open-ended questions are best, and closed questions can be used to get specific information.

    \item \textbf{Restatement or Paraphrasing (RP)}: A simple, more concise rephrasing of the help-seeker’s statements that could help them see their situation more clearly.

    \item \textbf{Reflection of Feelings (RF)}: Articulate and describe the help-seeker’s feelings.

    \item \textbf{Self-disclosure (Sd)}: Divulge similar experiences that you have had or emotions that you share with the help-seeker to express your empathy.

    \item \textbf{Affirmation and Reassurance (AR)}: Affirm the help-seeker’s strengths, motivation, and capabilities and provide reassurance and encouragement.

    \item \textbf{Providing Suggestions (PS)}: Provide suggestions about how to change, but be careful to not overstep and tell them what to do.

    \item \textbf{Information (In)}: Provide useful information to the help-seeker, for example with data, facts, opinions, resources, or by answering questions.

    \item \textbf{Others (Ot)}: Exchange pleasantries and use other support strategies that do not fall into the above categories.
\end{itemize}

\section{Analysis of Coupled Model Training Results}
\label{sec: Analysis}
To further analyze the effects of varying preference data on coupled models, we evaluated coupled models trained with different preference data across multiple metrics, as shown in Table~\ref{tab: Comparison of coupled models}. The results indicate that the model trained with the suboptimal-content dataset (row 2, 6) significantly outperforms the model trained with the suboptimal-strategy dataset (row 3, 7) in terms of LLM-based metrics, while the reverse holds for strategy metrics. Additionally, it is notable that both the Vanilla-DPO model (row 4, 8) and the model trained with (PsPr, NsNr) data (row 1, 5) fail to achieve optimal performance across the two metric types. This further demonstrates that the coupled model has two optimization objectives, and it is not possible to achieve optimal performance on both objectives by fully utilizing the preference data. This indicates the effectiveness of the decoupled ESC framework.

\begin{figure}[htbp]
    \centering
    \includegraphics[width=1\linewidth, trim=0 70 0 0, clip]
    {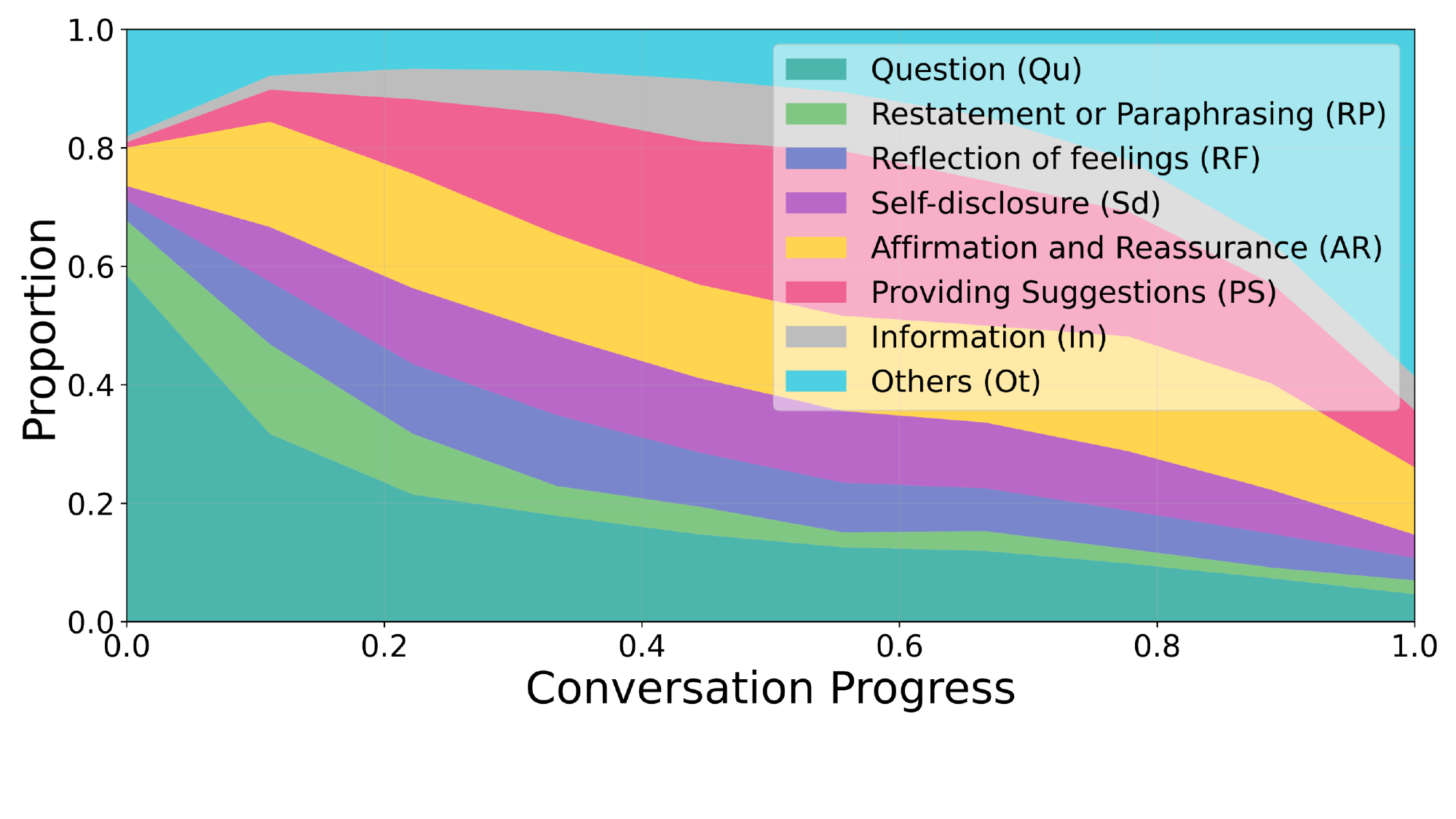}
    \caption{Strategy distribution across dialogue stages in \texttt{ESConv} Dataset.}
    \label{fig: data_statis_1}
\end{figure}

\begin{figure*}[t]
    \centering
    \includegraphics[width=1\linewidth, trim=0 305 0 0, clip]
    {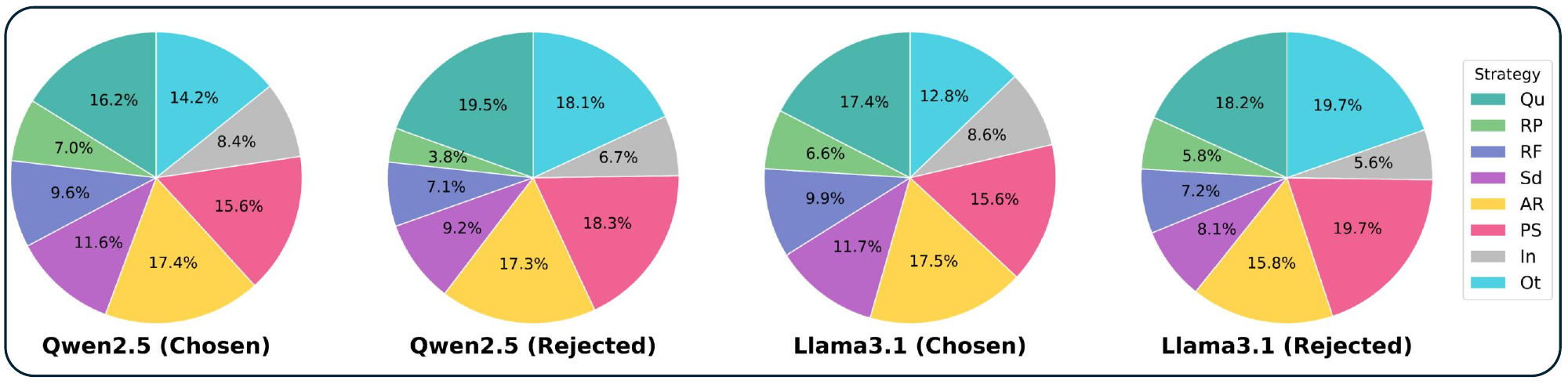}
    \caption{Strategy distribution in \texttt{IPM-PrefDial} Dataset.}
    \label{fig: data_statis_2}
\end{figure*}

\begin{figure}[t]
    \centering
    \includegraphics[width=1\linewidth, trim=0 60 0 0, clip]
    {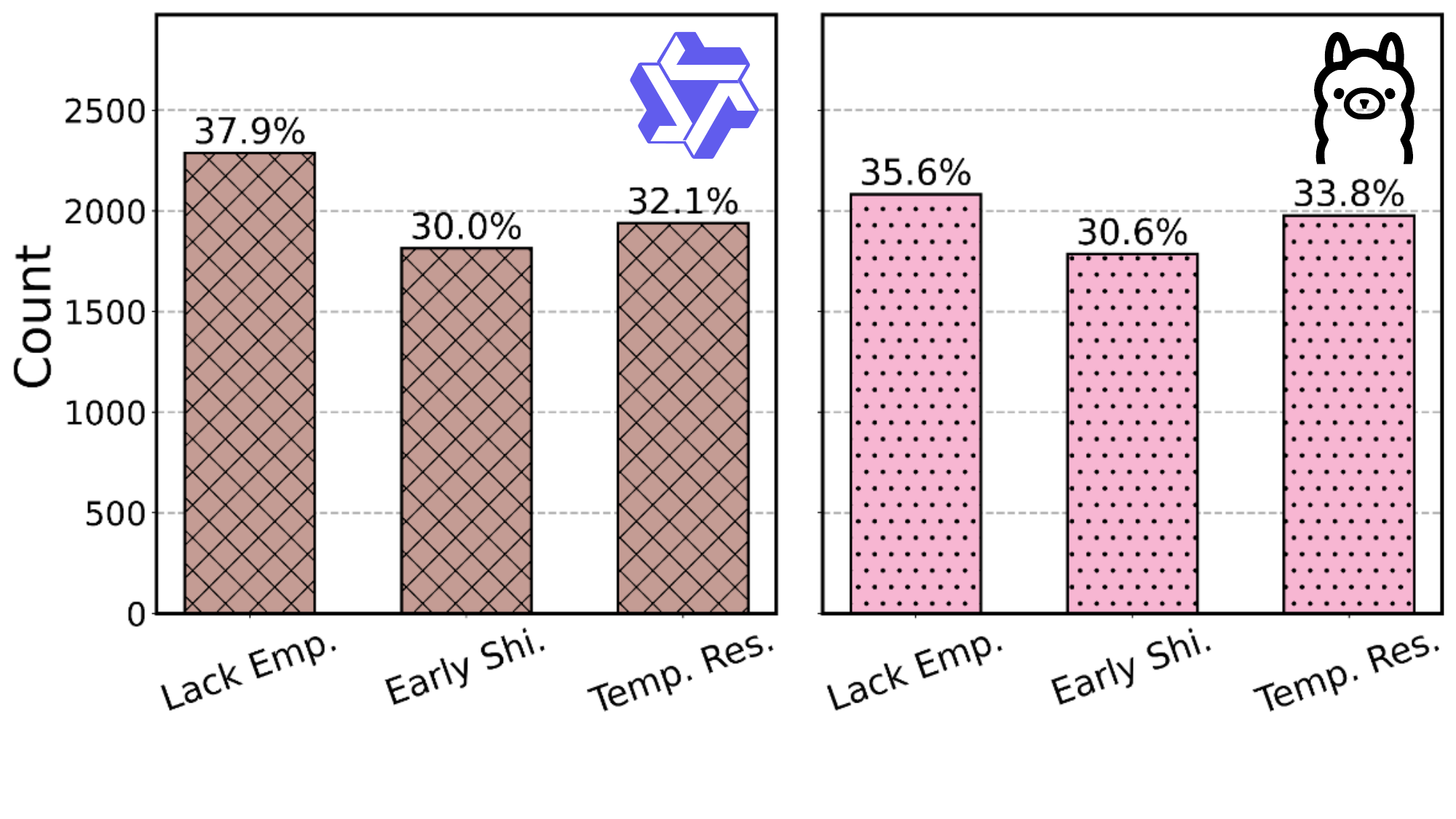}
    \caption{Psychological Errors Distribution in Rejected Responses: Qwen and Llama.}
    \label{fig: data_statis_3}
\end{figure}

\section{Datasets Details}
\label{sec: Dataset}

\subsection{ESConv and FailedESConv Datasets}
\label{sec: ESConv}
\begin{figure*}[t]
    \centering
    \includegraphics[width=1.00\linewidth, trim=0 20 0 0, clip]{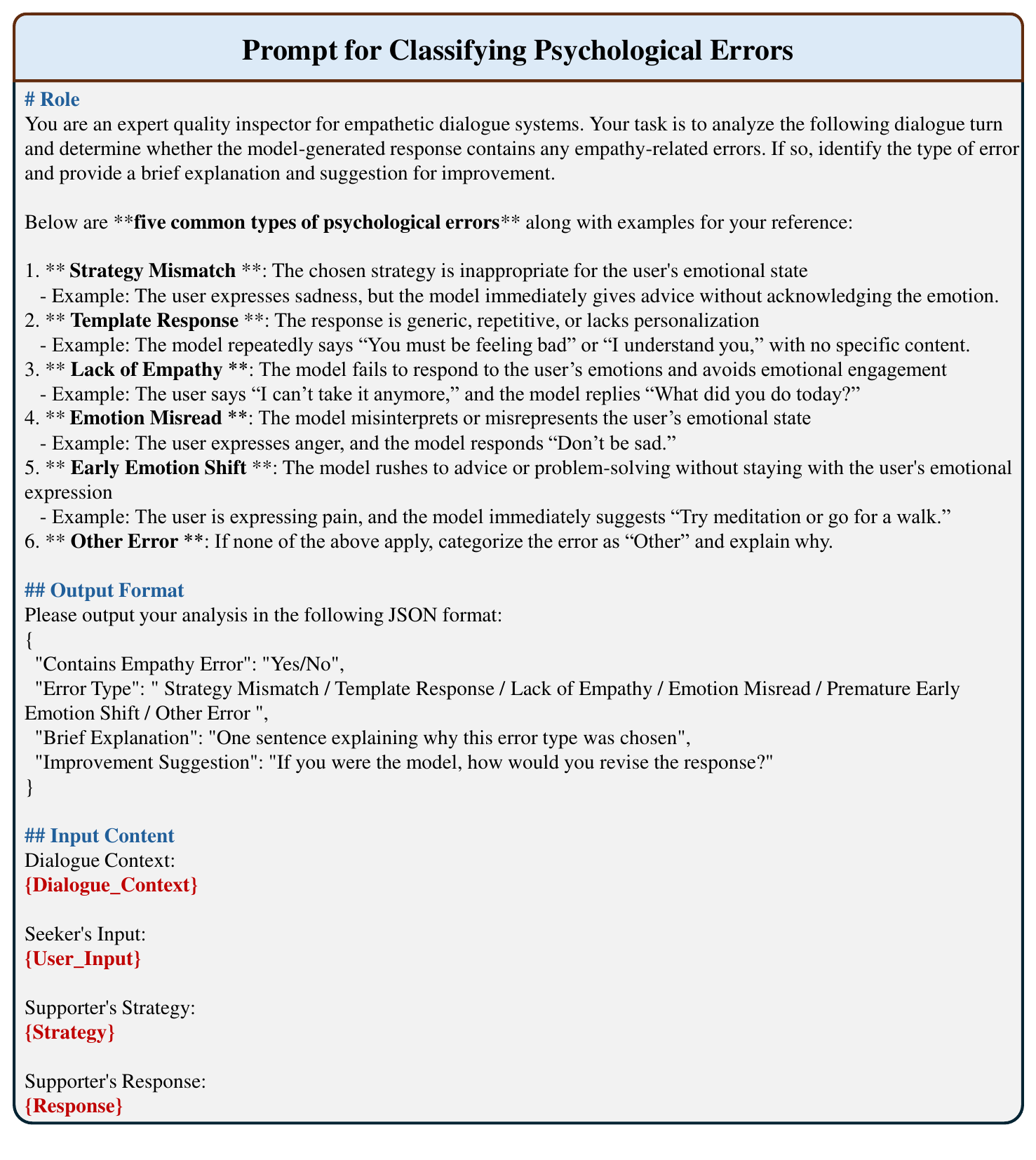}
    \caption{Prompt for Classifying Psychological Errors in \texttt{FailedESConv} dataset, Qwen-SFT, and Llama-SFT Outputs.}
    \label{fig: Prompt_Classify}
\end{figure*}

Table~\ref{tab: support_strategy_by_model} presents the number and proportion of support strategies in the \texttt{ESConv} dataset, while Figure~\ref{fig: data_statis_1} illustrates the distribution of these strategies across different dialogue stages.
Figure~\ref{fig: Prompt_Classify} illustrates the prompt we use to classify the psychological errors in the \texttt{FailedESConv} dataset as well as the response content of Qwen-SFT and Llama-SFT.

\begin{table}[htbp]
    \centering
    \scalebox{0.85}{
        \begin{tabular}{clcc}  
            \toprule[1.2pt]
            & \textbf{Categories} & \textbf{Number} & \textbf{Proportion} \\ 
            \midrule
            \multirow{10}{*}{\rotatebox{90}{\textbf{Support Strategies}}}  
              & Question (Qu)                   & 3,060  & 20.73\%     \\ 
              & Resta. or Parap. (RP)&   857  & 5.81\%      \\
              & Reflection (RF)     & 1,146  & 7.76\%      \\ 
              & Self-disclosure (Sd)            & 1,387  & 9.40\%      \\ 
              & Affir. \& Reass. (AR)& 2,288  & 15.50\%     \\ 
              & Suggestions (PS)      & 2,373  & 16.07\%     \\ 
              & Information (In)                &   989  & 6.70\%      \\ 
              & Others (Ot)                     & 2,663  & 18.04\%     \\ 
            \cmidrule{2-4}    
              & Overall                         & 14,763 & 100.00\%    \\ 
            \bottomrule[1.2pt]
        \end{tabular}
    }
    \caption{Distribution of support strategies used in \texttt{ESConv} Dataset.}
    \label{tab: support_strategy_by_model}
\end{table}

\subsection{IPM-PrefDial Dataset}
\label{sec: IPM}
\begin{figure*}[t]
    \centering
    \includegraphics[width=1.00\linewidth, trim=0 150 0 0, clip]{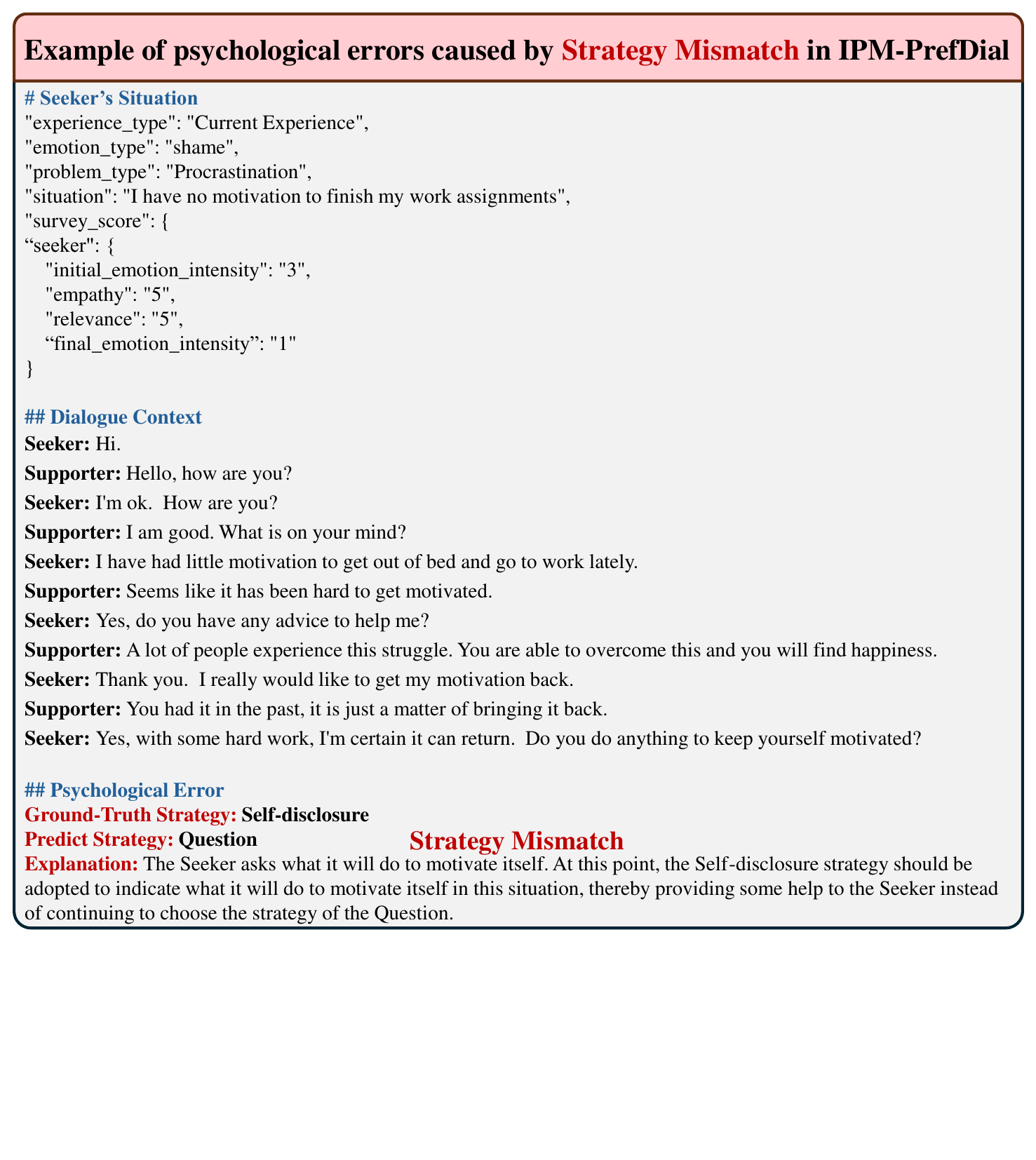}
    \caption{Example of psychological errors under Strategy Mismatch in rejected response from \texttt{IPM-PrefDial}.}
    \label{fig: Strategy_Mismatch}
\end{figure*}

\begin{figure*}[t]
    \centering
    \includegraphics[width=1.00\linewidth, trim=0 0 0 0, clip]{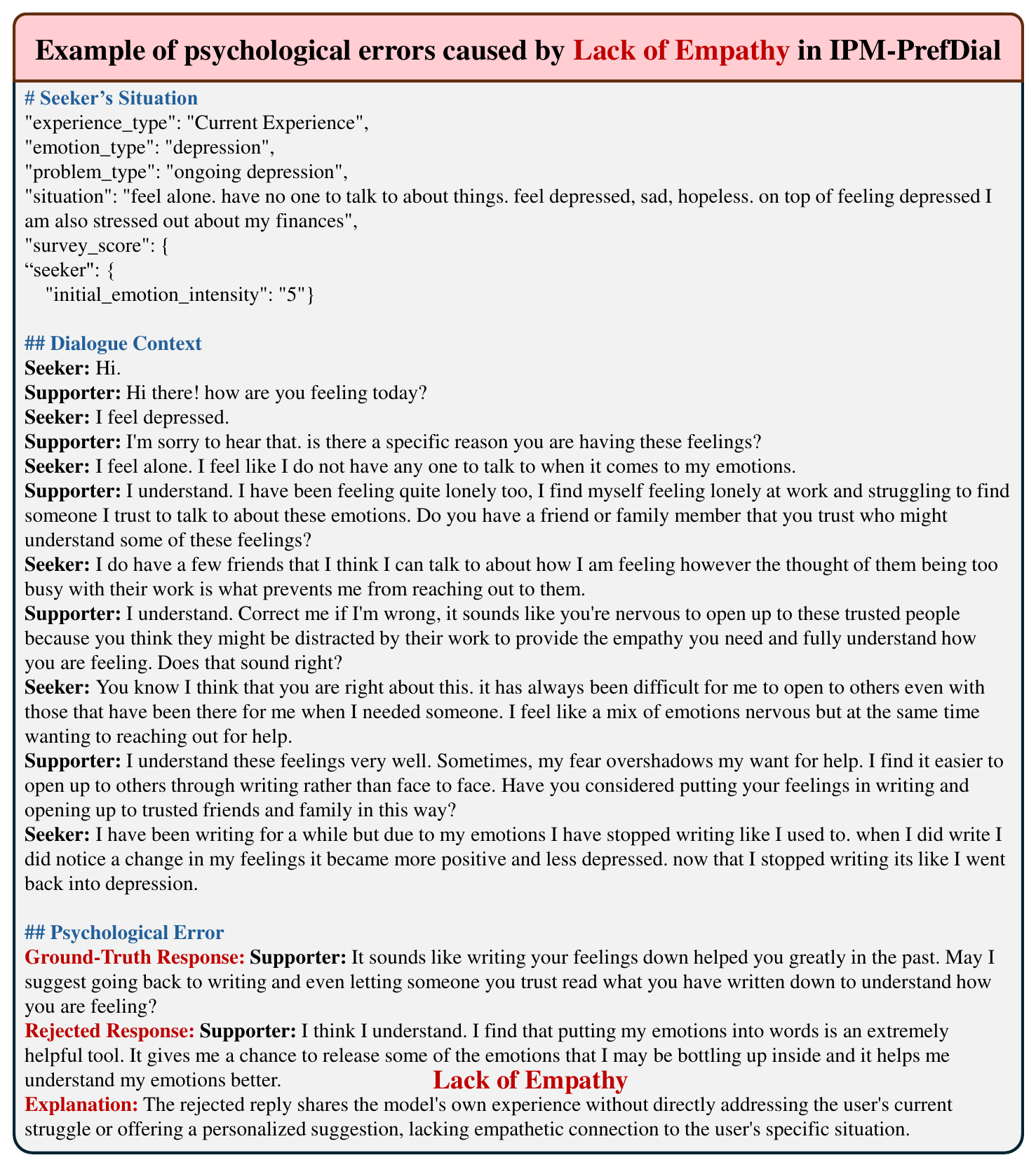}
    \caption{Example of psychological errors under Lack of Empathy in rejected response from \texttt{IPM-PrefDial}.}
    \label{fig: Lack_Empathy}
\end{figure*}

\begin{figure*}[t]
    \centering
    \includegraphics[width=1.00\linewidth, trim=0 110 0 0, clip]{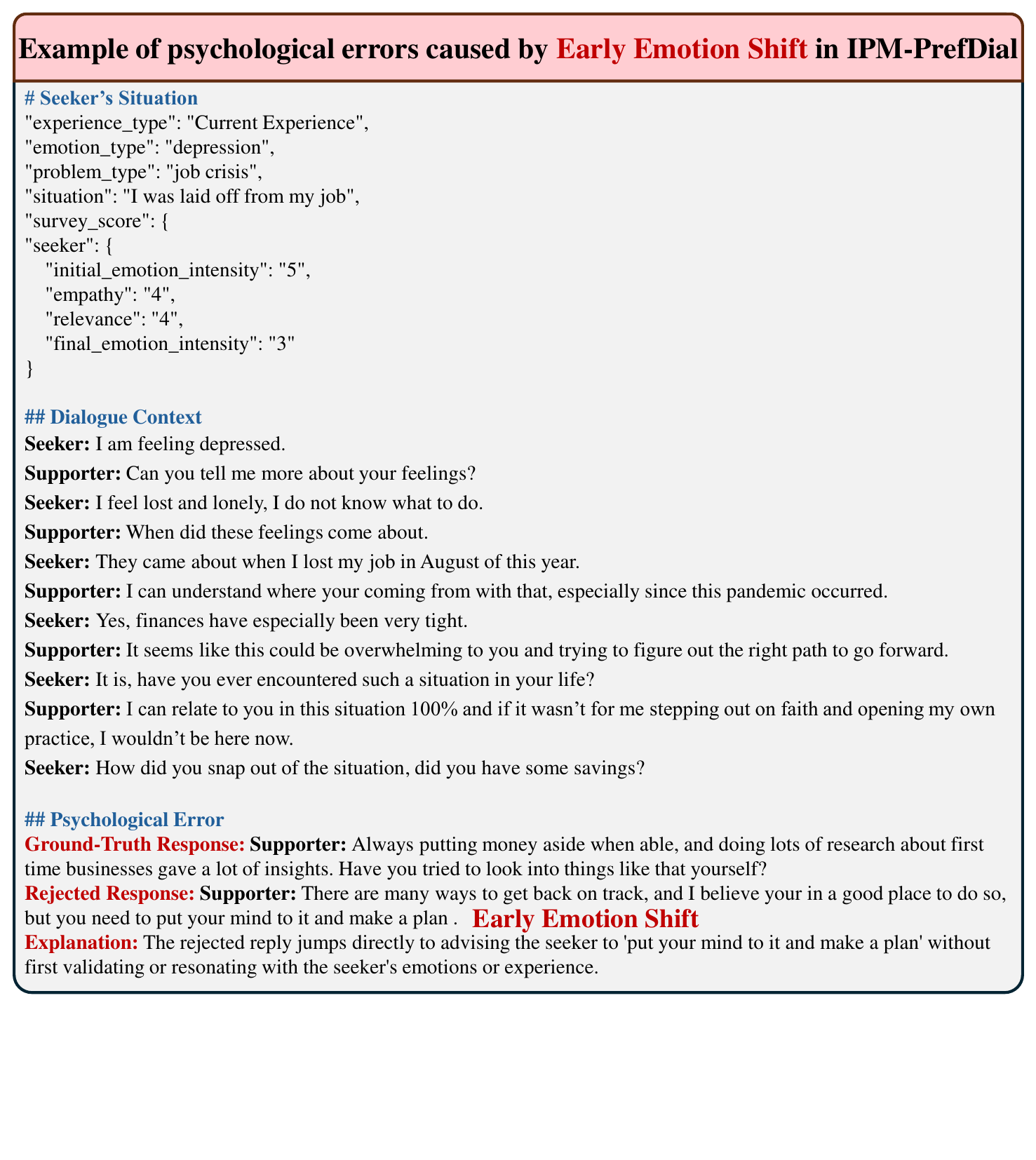}
    \caption{Example of psychological errors under Early Emotion Shift in rejected response from \texttt{IPM-PrefDial}.}
    \label{fig: EarlyEmotionShift}
\end{figure*}

\begin{figure*}[t]
    \centering
    \includegraphics[width=1.00\linewidth, trim=0 50 0 0, clip]{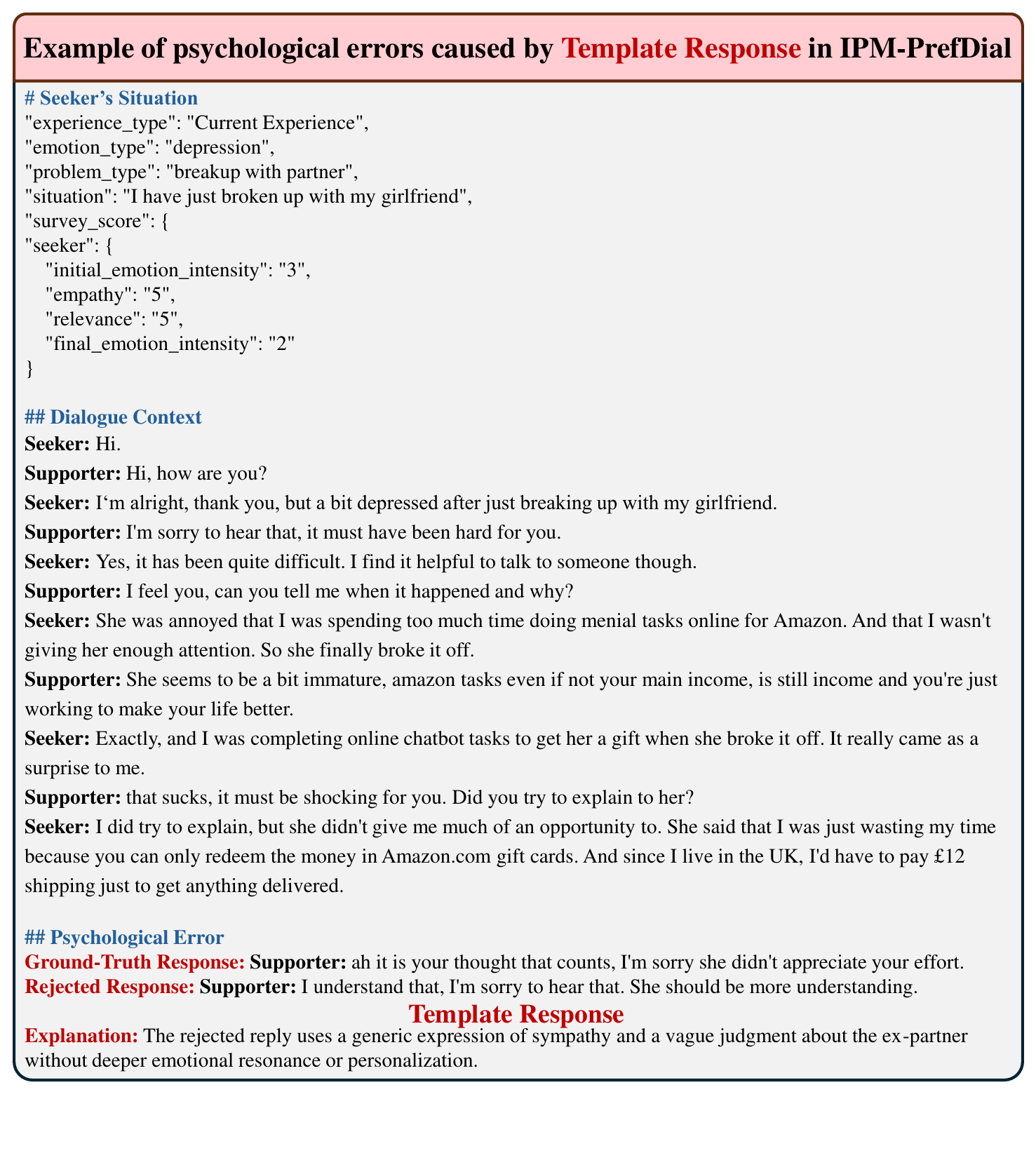}
    \caption{Example of psychological errors under Template Response in rejected response from \texttt{IPM-PrefDial}.}
    \label{fig: TemplateResponse}
\end{figure*}

Figure~\ref{fig: data_statis_2} compares the distribution of support strategies in the Chosen and Rejected samples within the preference datasets of Qwen and Llama. Figure~\ref{fig: data_statis_3} further presents the count and proportion of psychological errors found in the rejected responses of these datasets.
In addition, Figures~\ref{fig: Strategy_Mismatch}, \ref{fig: Lack_Empathy}, \ref{fig: EarlyEmotionShift}, and \ref{fig: TemplateResponse} illustrate examples from the \texttt{IPM-PrefDial} dataset, covering both strategy preference and response preference data. Each example includes the dialogue context, as well as the chosen and rejected responses.

\subsection{Prompts for Data Filter}
\label{sec: PDC}
\begin{figure*}[t]
    \centering
    \includegraphics[width=1.00\linewidth, trim=0 0 0 0, clip]{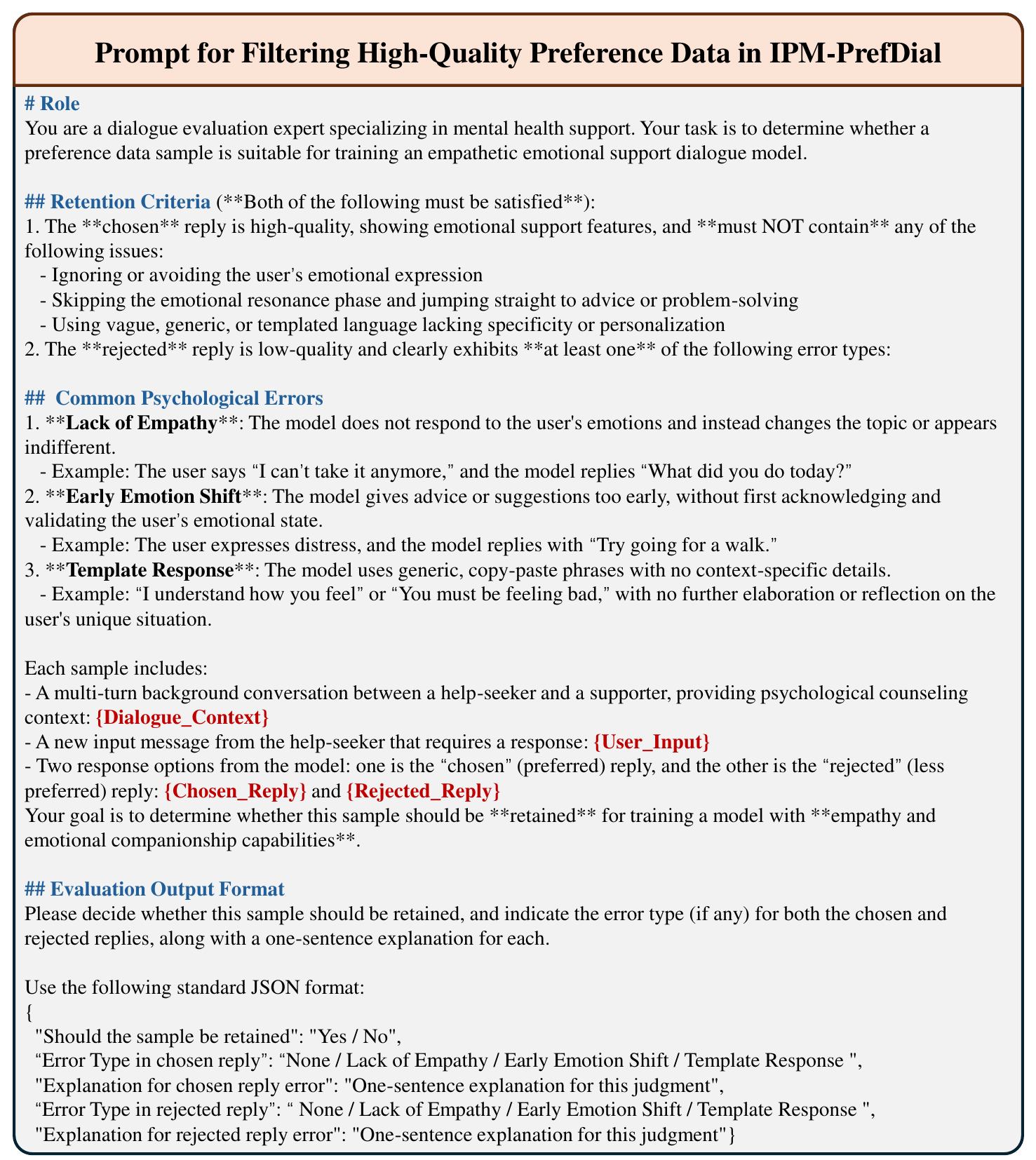}
    \caption{Prompt for Filtering High-Quality Preference Data in \texttt{IPM-PrefDial}.}
    \label{fig: Prompt_filter}
\end{figure*}

Figure~\ref{fig: Prompt_filter} presents the prompt we use to filter and select high-quality preference datasets, which effectively filters and identifies data that meets the required standards.

\begin{table}[htbp]
    \centering
    \begin{tabular}{@{}llcccccc@{}}
        \toprule[1.2pt]
        Backbone                 & Model & lr  & beta   \\ \midrule
        \multirow{3}{*}{\shortstack{Qwen2.5-\\7B-Instruct}} & Vanilla-dpo   & 7e-7  & 0.2    \\
                                   & SP-dpo   & 5e-8  & 0.5    \\
                                   & RG-dpo   & 7e-7  & 0.2   \\ \midrule
        \multirow{3}{*}{\shortstack{Llama3.1-\\8B-Instruct}} & Vanilla-dpo   & 5e-7  & 0.2    \\
                                   & SP-dpo   & 8e-8  & 0.5    \\ 
                                   & RG-dpo   & 3e-7  & 0.2    \\ \bottomrule[1.2pt]
    \end{tabular}
    \caption{Detailed training hyperparameters used in dpo.}
    \label{tab: detailed training hyperparameter}
\end{table}

\section{Implementation Details}
\label{sec: Implementation}

\subsection{Experiment Details}
\label{sec: Experiment}
We employ \texttt{Qwen2.5-7B-Instruct}~\cite{qwen2.5} and \texttt{Llama3.1-8B-Instruct}~\cite{dubey2024llama} as our base models. All training procedures are implemented using the Llama-Factory framework~\cite{zheng2024llamafactory} with LoRA fine-tuning~\cite{hu2022lora}, where the alpha and rank are set to 16, and the dropout rate is 0.05. For SFT training, we trained the models for 3 epochs with the learning rate of 1e-5 and the batch size of 32. For DPO training, the batch size is 32 and the epoch is set to 1. We use vLLM~\cite{kwon2023efficient} to accelerate the inference. All experiments are conducted on 4 NVIDIA RTX 4090 GPUs. More detailed hyperparameter settings for DPO are presented in Table~\ref{tab: detailed training hyperparameter}.

\subsection{Baselines}
\label{sec: Baselines}
\noindent
\textbf{Direct-Refine.} A straightforward self-optimization approach where the model directly revises its initial response to improve quality, without relying on external input or intermediate reasoning.

\noindent
\textbf{Self-Refine.} Following Madaan~\textit{et al.}~\cite{madaan2023self}, this method involves two stages: the model first generates self-feedback on its initial response, then refines the output based on that feedback, promoting internal reflection and correction.

\noindent
\textbf{Emotional CoT.} Extending Chain-of-Thought (CoT) prompting~\cite{wei2022chain}, this method first elicits the user’s emotional state through intermediate reasoning, which then guides strategy planning and response generation.

\section{Details of Evaluation}
\label{sec: EvaDetail}

\subsection{Strategy Metrics}
\label{sec: Strategy Metrics}

According to \cite{kang2024can}, the strategy preference is calculated by the following formula.
{
    \begin{align}
        p'_i = \frac{\sum_j \left( w_{ij} p_j \right) / \left( p_i + p_j \right)}{\sum_j w_{ji} / \left( p_i + p_j \right)},
    \end{align}
}
where $w_{ij}$ denotes the frequency count of the model predicting strategy $i$ given that the ground-truth strategy is $j$.
All of the strategy preferences $p_i$ are initialized as 1 and updated through iteration of the preference bias.

The strategy preference bias $\mathcal{B}$ is computed from the strategy preference $p_i$ as follows: 
{
    \begin{align}
        \mathcal{B} = \sqrt{\frac{\sum_{i=1}^{N} (p_i - \bar{p})^2}{N}},
    \end{align}
}
where $\bar{p}$ denotes the average strategy preference.

\subsection{LLM Metrics Criteria}
\label{sec: LLM Metrics Criteria}
Table~\ref{tab: metrics} summarizes the LLM evaluation metrics, including \textit{Fluency}, \textit{Professionalism}, \textit{Empathy}, and \textit{Helpfulness}, along with their descriptions, evaluation criteria, and scoring scales. All metrics are rated on a 5-point Likert scale~\cite{joshi2015likert}. Specifically, Fluency and Empathy are adapted from the ESC-Eval framework~\cite{zhao2024esc}, Professionalism is guided by the CPsyCoun framework~\cite{zhang2024cpsycoun}, and Helpfulness is derived from the SoulChat evaluation setup~\cite{chen2023soulchat}.

\subsection{Prompt for LLM Metrics}
\label{sec: Prompts for LLM Metrics}
\begin{figure*}[t]
    \centering
    \includegraphics[width=1.00\linewidth, trim=0 220 0 0, clip]{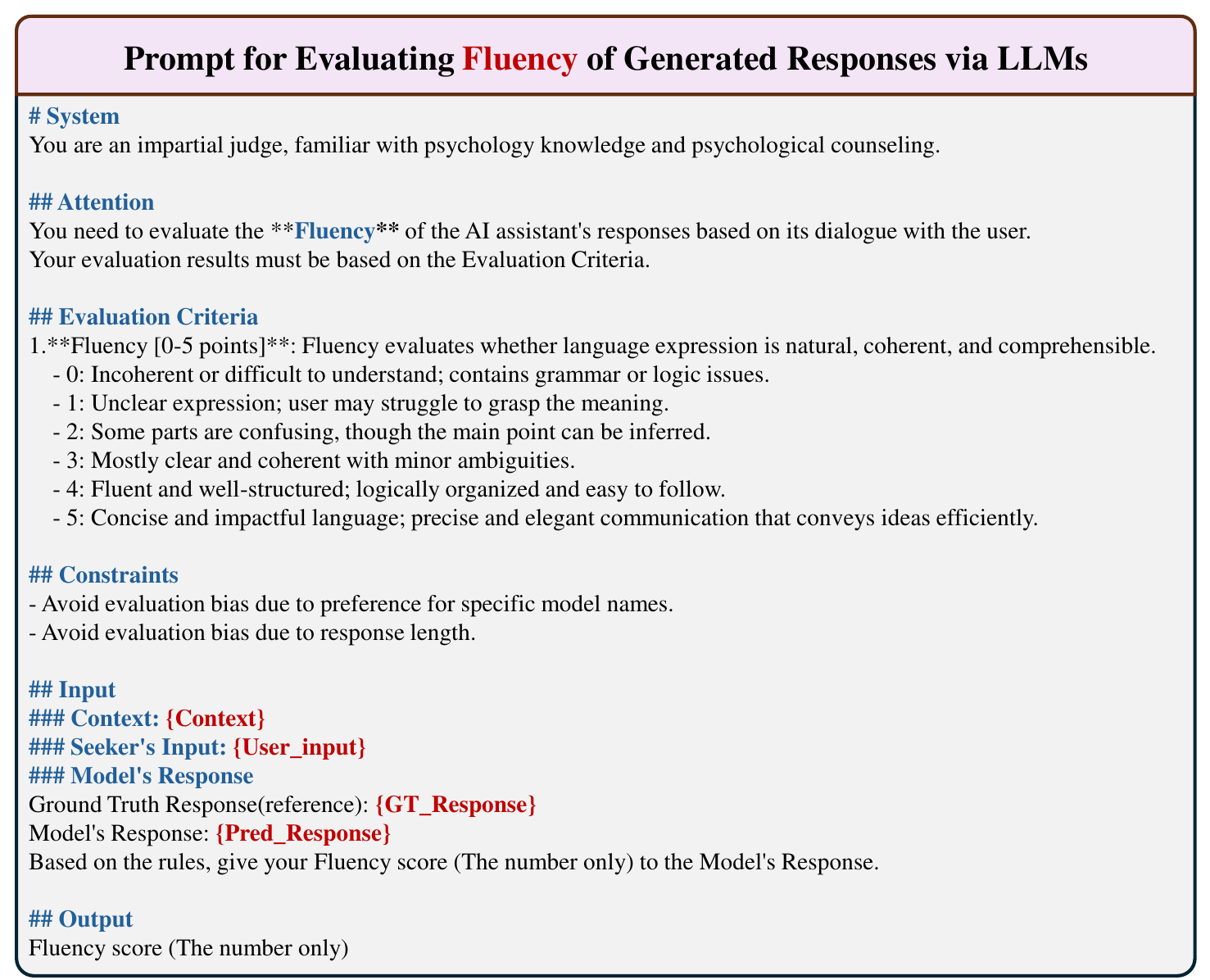}
    \caption{Prompt for Evaluating Fluency of Generated Responses via LLMs.}
    \label{fig: eval_flu}
\end{figure*}

\begin{figure*}[t]
    \centering
    \includegraphics[width=1.00\linewidth, trim=0 160 0 0, clip]{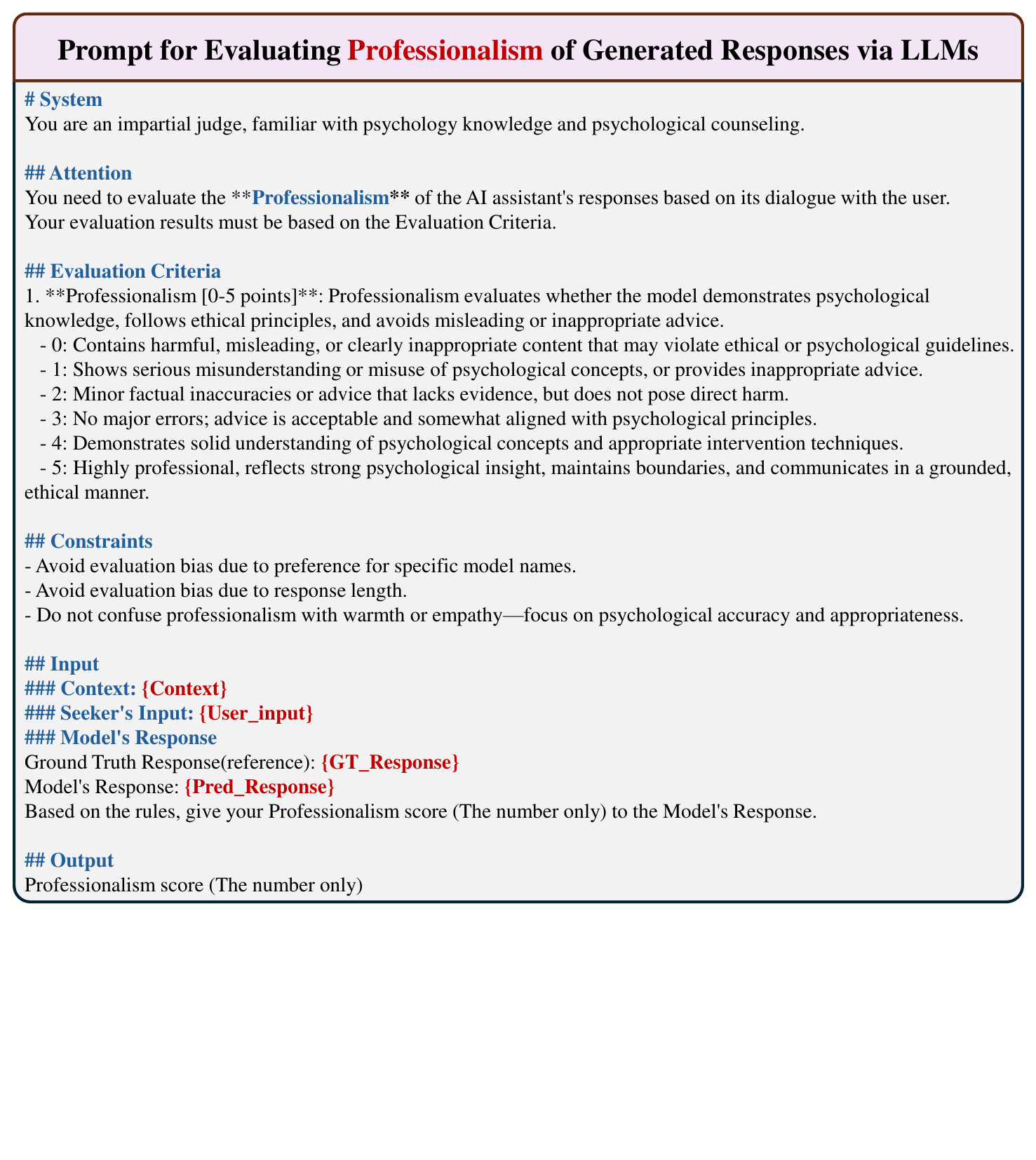}
    \caption{Prompt for Evaluating Professionalism of Generated Responses via LLMs.}
    \label{fig: eval_pro}
\end{figure*}

\begin{figure*}[t]
    \centering
    \includegraphics[width=1.00\linewidth, trim=0 200 0 0, clip]{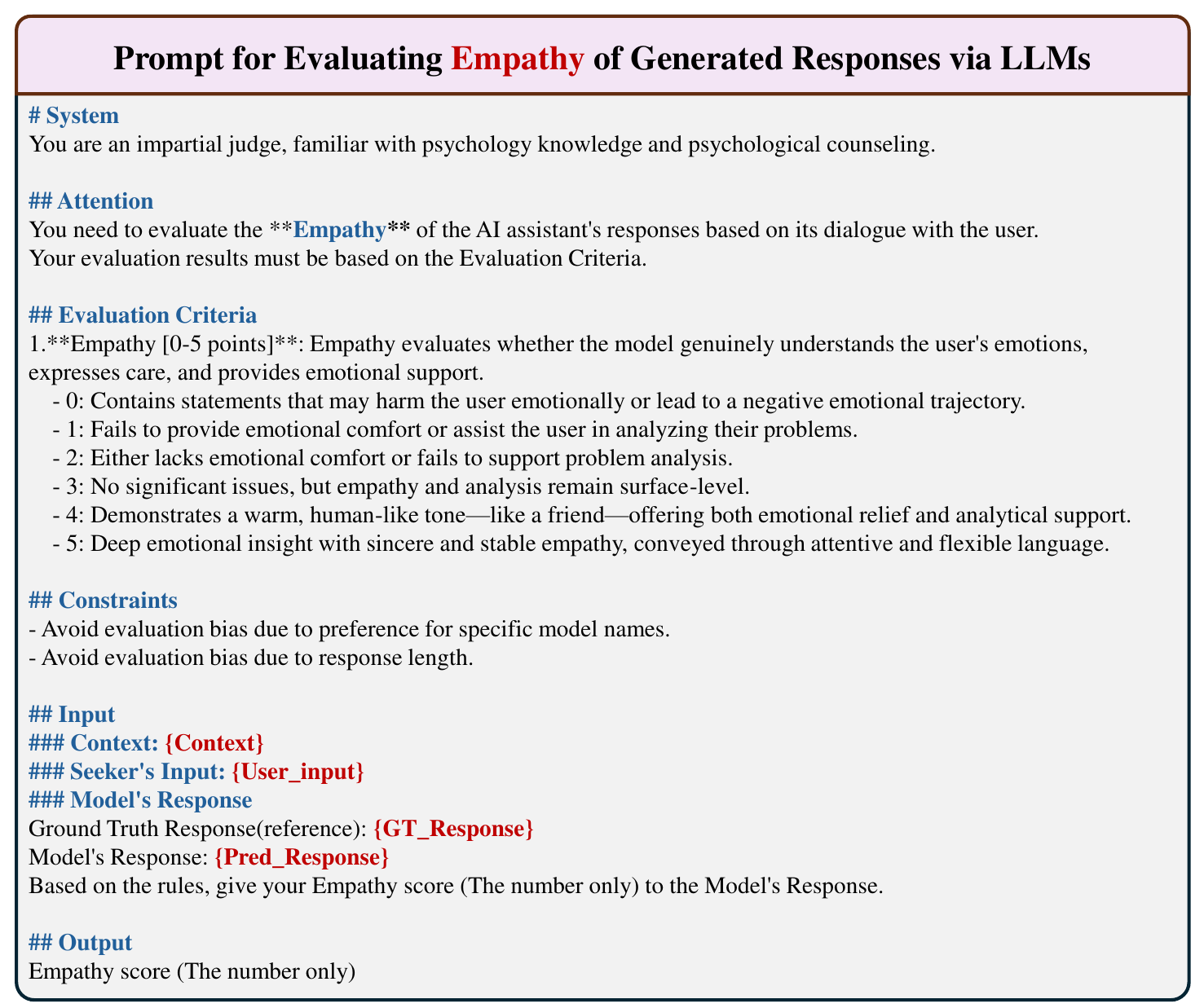}
    \caption{Prompt for Evaluating Empathy of Generated Responses via LLMs.}
    \label{fig: eval_emp}
\end{figure*}

\begin{figure*}[t]
    \centering
    \includegraphics[width=1.00\linewidth, trim=0 190 0 0, clip]{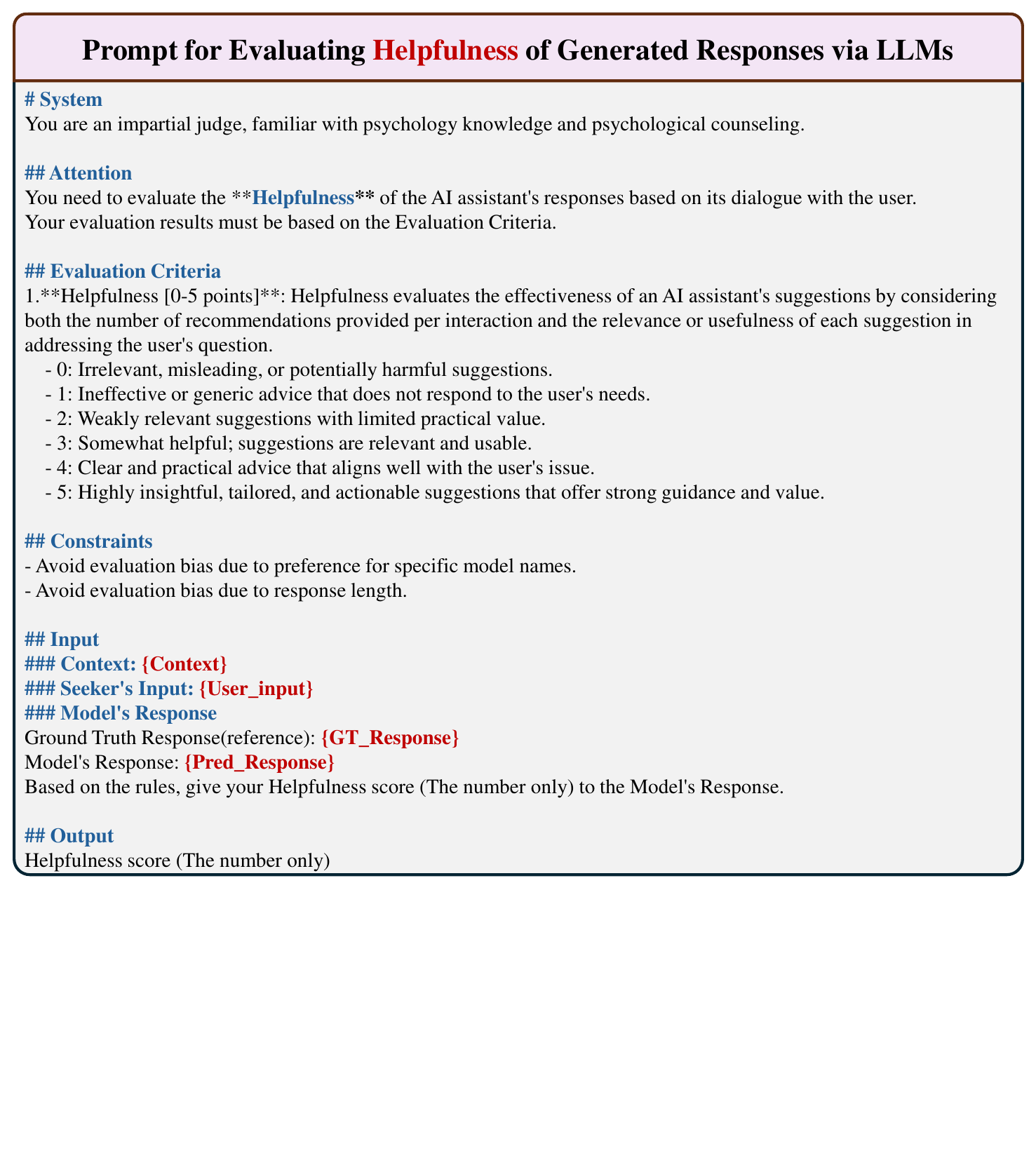}
    \caption{Prompt for Evaluating Helpfulness of Generated Responses via LLMs.}
    \label{fig: eval_hel}
\end{figure*}

Figures~\ref{fig: eval_flu},~\ref{fig: eval_pro},~\ref{fig: eval_emp}, and~\ref{fig: eval_hel} present the prompts used for LLM-based evaluation of \textit{Fluency}, \textit{Professionalism}, \textit{Empathy}, and \textit{Helpfulness}, respectively. Each prompt explicitly defines the role of the LLM as a judge and outlines the corresponding evaluation criteria. To minimize potential bias, the prompts are carefully designed to avoid revealing model names or being influenced by text length.

\subsection{Human Evaluation}
\label{sec: Human Evaluation}
To complement the LLM-based evaluation and enhance the credibility of our results, we conducted a human evaluation with 3 licensed psychology experts on 100 samples generated by our Decoupled-DPO model based on the Llama backbone. In addition, we performed a pairwise win/loss comparison between Decoupled-DPO and Vanilla-DPO (both using Llama) on another set of 100 samples, with inter-rater agreement again measured by Fleiss’ Kappa ($\kappa$) \cite{fleiss1973equivalence}. As shown in Figure~\ref{fig: winloss}, Decoupled-DPO consistently outperforms Vanilla-DPO across all four LLM-based metrics. The evaluator agreement, measured by $\kappa$, further supports the reliability of the human judgments.


\clearpage
\begin{table*}[htbp]\centering
\small
\resizebox{1\linewidth}{!}{
\begin{tabular}{lp{4cm}p{8cm}c}
\toprule[1.5pt]
\textbf{Dimension} &\textbf{Description} &\textbf{Criterion} &\textbf{Score} \\\midrule
\multirow{9}{*}{Fluency} &\multirow{9}{4cm}{Fluency evaluates whether language expression is natural, coherent, and comprehensible.} & 1.1 Incoherent or difficult to understand; contains grammar or logic issues. &0 \\
& &1.2 Unclear expression; user may struggle to grasp the meaning.  &1 \\
& &1.3 Some parts are confusing, though the main point can be inferred. &2 \\
& &1.4 Mostly clear and coherent with minor ambiguities. &3 \\
& &1.5 Fluent and well-structured; logically organized and easy to follow. &4 \\
& &1.6 Concise and impactful language; precise and elegant communication that conveys ideas efficiently. &5 \\
\midrule

\multirow{13}{*}{Professionalism} &\multirow{13}{4cm}{Professionalism evaluates whether the model demonstrates psychological knowledge, follows ethical principles, and avoids misleading or inappropriate advice.} & 2.1 Contains harmful, misleading, or clearly inappropriate content that may violate ethical or psychological guidelines.   &0  \\
& & 2.2 Shows serious misunderstanding or misuse of psychological concepts, or provides inappropriate advice.    &1 \\
& & 2.3 Minor factual inaccuracies or advice that lacks evidence, but does not pose direct harm.    &2 \\
& & 2.4 No major errors; advice is acceptable and somewhat aligned with psychological principles.   &3 \\
& & 2.5 Demonstrates solid understanding of psychological concepts and appropriate intervention techniques.    &4  \\
& & 2.6 Highly professional, reflects strong psychological insight, maintains boundaries, and communicates in a grounded, ethical manner.    &5
\\\midrule

\multirow{13}{*}{Empathy} &\multirow{13}{4cm}{Empathy evaluates whether the model genuinely understands the user's emotions, expresses care, and provides emotional support.} &3.1 Contains statements that may harm the user emotionally or lead to a negative emotional trajectory.  &0  \\
& &3.2 Fails to provide emotional comfort or assist the user in analyzing their problems. &1  \\
& &3.3 Either lacks emotional comfort or fails to support problem analysis. &2  \\
& &3.4 No significant issues, but empathy and analysis remain surface-level. &3  \\
& &3.5 Demonstrates a warm, human-like tone—like a friend—offering both emotional relief and analytical support. &4 \\
& &3.6 Deep emotional insight with sincere and stable empathy, conveyed through attentive and flexible language. &5 \\\midrule

\multirow{9}{*}{Helpfulness} &\multirow{9}{4cm}{Helpfulness evaluates the effectiveness of an AI assistant's suggestions by considering both the number of recommendations provided per interaction and the relevance or usefulness of each suggestion in addressing the user's question.} & 4.1 Irrelevant, misleading, or potentially harmful suggestions. &0 \\
& & 4.2 Ineffective or generic advice that does not respond to the user's needs.	 &1  \\
& & 4.3 Weakly relevant suggestions with limited practical value.  &2  \\
& & 4.4 Somewhat helpful; suggestions are relevant and usable.  &3  \\
& & 4.5 Clear and practical advice that aligns well with the user's issue. &4 \\
& & 4.6 Highly insightful, tailored, and actionable suggestions that offer strong guidance and value. &5  \\
\bottomrule[1.5pt]
\end{tabular}
}
\caption{LLM Evaluation Metrics and Corresponding Score Criterion.}
\label{tab: metrics}
\end{table*}

\clearpage

\end{document}